\documentclass[conference]{IEEEtran}

\usepackage{graphicx}
\usepackage{amsmath,amssymb}
\usepackage{booktabs}
\usepackage{multirow}
\usepackage{enumitem}
\usepackage{cite}
\usepackage{hyperref}

\title{A Survey on Sensor-based Planning and Control for Unmanned Underwater Vehicles}

\author{%
\IEEEauthorblockN{Shivam Vishwakarma, Tejal Bedmutha, Dharmendra Kumar Patel, Vijay Bhaskar Semwal, Leena Vachhani}
\IEEEauthorblockA{%
IIT Bombay, Mumbai, India \\
Aerospace Engineering, IIT Bombay, Mumbai, India \\
CAIR, DRDO, Bengaluru, India \\
MANIT Bhopal, India
}
}
\begin{document}
\maketitle
\begin{abstract}
This survey examines recent sensor-based planning and control methods for Unmanned Underwater Vehicles (UUVs). In complex, uncertain underwater environments, UUVs require advanced planning and control strategies for effective navigation. These vehicles face significant challenges including drifting and noisy sensor measurements, absence of Global Navigation Satellite System (GNSS) signals, and low-bandwidth, high-latency underwater acoustic communications. The focus is on reactive local planning layers that adapt
to real-time sensor inputs such as SONAR and Inertial Measurement Units (IMU) to improve localization accuracy and autonomy in dynamic ocean conditions, enabling dynamic obstacle avoidance and on-the-fly re-planning.
The survey categorizes the existing literature into decoupled and coupled architectures for sensor-based planning and control. The decoupled architecture sequentially addresses planning and control stages, whereas coupled architectures offer tighter feedback loops for more immediate responsiveness. A comparative analysis of coupled planning and control methods reveals that while PID controllers are simple, they lack predictive capability for complex maneuvers. Model Predictive Control (MPC) offers superior path optimization but can be computationally intensive, and invariant-set controllers provide strong safety guarantees at the potential cost of agility in confined environments. Key
contributions include a taxonomy of architectures combining planning and control, a focus on adaptive local planning, and an analysis of controller roles in integrated planning frameworks for autonomous navigation of UUVs.
\end{abstract}


\vspace{6pt}
\noindent\textbf{Abbreviations:} UUVs, Unmanned Underwater Vehicles; PID, Proportional Integral Derivative; MPC, Model Predictive Control; IPC, Integrated Planning and Control; IMU, Inertial Measurement Units.

\renewcommand\thefootnote{\fnsymbol{footnote}}
\setcounter{footnote}{1}

\section{Introduction}\label{sec1}

Unmanned Underwater Vehicles (UUVs) are a cornerstone of modern maritime operations, enabling automated exploration, inspection, surveillance and environmental monitoring in oceanic domains that are often unreachable or inhospitable to human divers. Their autonomy, which is their capability to perceive, decide and act independently without continuous human intervention, has revolutionized the way we interact with sub-sea environments, enabling vehicles to navigate, adapt and perform complex tasks even under uncertain and dynamic conditions. This self-directed behavior supports missions ranging from deep-sea resource exploration and coral reef monitoring to underwater infrastructure inspection, mine detection and search-and-rescue operations in complex, unstructured domains. (\cite{Leonard2013, Miller2010, Williams2001}). To better illustrate the diversity of UUV applications, Table ~\ref{tab:uuv_applications} summarizes common operational domains and their associated objectives.

\begin{table*}[ht]
\centering
\caption{Applications of UUVs: }
\renewcommand{\arraystretch}{1.3}
\begin{tabular}{@{}p{4cm}p{5.5cm}p{6.5cm}@{}}
\toprule
\textbf{Application Domain} & \textbf{Description} & \textbf{Corresponding academic reference} \\
\midrule

Deep-Sea Exploration & Energy-conserving, collision-free path planning in unknown marine environments & 
(\cite{Danovaro2014}), 
(\cite{Meng2025})
\\

Environmental Monitoring & Continuously tracking environmental conditions to assess and protect global ecosystems & 
(\cite{Artiola2004}),  
(\cite{OKEREKE2023})

\\
Military Operations & Securing seas, projecting power, defending national interests & 
(\cite{Livingston2002}), 
(\cite{Ma2023}) 
\\

Underwater Inspection & Underwater drones navigate using diverse sensors for extended, pilot-less missions & 
(\cite{Hollinger2013}), 
(\cite{Behnaz2022})
\\

Scientific Research &  Advanced tech for complex underwater exploration and mapping & 
(\cite{Martin2016}), 
(\cite{Hasan2024})
\\

\bottomrule
\end{tabular}
\label{tab:uuv_applications}
\end{table*}

At the heart of this autonomy lies a suite of tightly integrated architecture that brings together autonomous navigation, sensor fusion and mechanical design optimization. Central to autonomous navigation are the core modules of mapping, localization, path planning and motion control. Mapping plays a pivotal role in autonomous underwater navigation by transforming raw sensor data into structured representations of the surrounding environment. It serves as a critical intermediary between perception and navigation, facilitating both situational awareness and localization. Through data obtained from SONAR, Doppler Velocity Logs (DVL), Inertial Measurement Units (IMU) and sometimes optical sensors, mapping algorithms generate spatial models such as occupancy grids, bathymetry maps or point clouds that describe environmental features like obstacles, terrain contours and free space 
helps in sensing the environment and localizing, respectively. Furthermore, the situational awareness derived from mapping forms the foundation upon which path planning algorithms generating safe, feasible and efficient trajectories from a start position to a target, considering the constraints of the environment and the vehicle’s dynamics (\cite{Bobkov2018, Jalal2021}). It also informs the control module by ensuring that trajectory tracking remains consistent with the actual environment. While maintaining stability and resilience in the face of disturbances. In UUVs navigation, planning, and control are not just routine functions—they are mission-critical elements whose performance directly dictates the operation's safety, efficiency and success. 


\begin{figure*}[ht]
    \centering
    \includegraphics[width=0.9\textwidth]{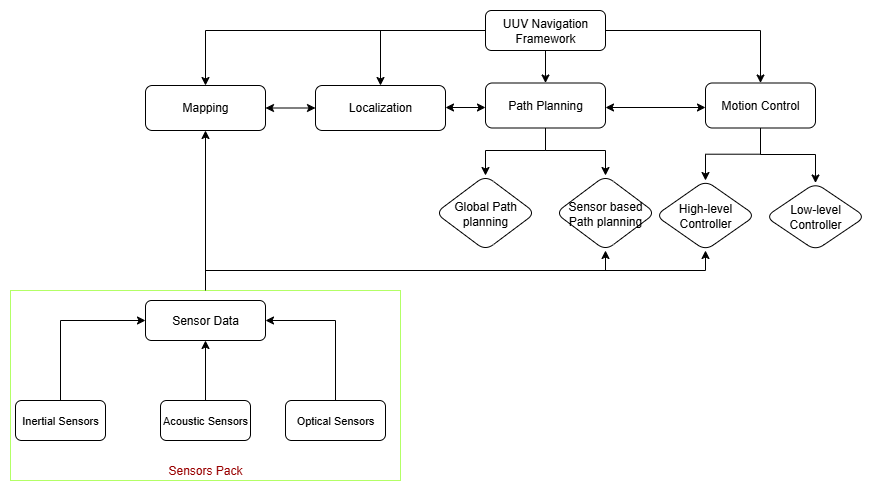}
    \caption{ UUV Navigation Stack Hierarchy}
    \label{fig:uuv_hier}
\end{figure*}

Additionally, the underwater environment presents limited visibility and perception fidelity. Optical cameras suffer from turbidity, light attenuation and scattering, especially in deep or murky waters. As a result, acoustic sensors such as side-scan sonar, multi-beam echo-sounder and imaging sonar are the primary tools for perception (\cite{Elkins2010, Chang2022}). However, these sensors provide relatively sparse and noisy data, making it difficult to distinguish between terrain features and dynamic obstacles in real time (\cite{Panda2020}). Another challenge stems from the dynamic and non-uniform nature of underwater forces. Ocean currents, thermoclines, salinity gradients and waves all introduce continuous disturbances that can alter a UUVs path and orientation (\cite{Sahoo2022, Orłowski2022, Shetty2021}). These hydrodynamic influences vary with depth, location and time and are often unknown or unpredictable in advance. Moreover, physical interactions such as tether forces in Remotely Operated Vehicles (ROVs) or changes in buoyancy add additional complexity to the control system (\cite{Yin2021, Petillot1998}). The communication limitations are equally significant. Unlike aerial or ground robots that can rely on high-bandwidth radio communication, UUVs communicate primarily through acoustic modems, characterized by low data rates, significant latency and susceptibility to noise and interference. This imposes a strong requirement for autonomy; once submerged, UUVs must independently make planning and control decisions onboard, without relying on real-time operator input (\cite{Yilmaz2008, Li2019, Zeng2015}). Given these constraints, planning strategies must address two fundamental aspects: long-range mission objectives and immediate reactive responses. This dichotomy is reflected in the concepts of global and sensor-based path planning (\cite{RONGHAO2025, Wynn2014}). Global planners design high-level trajectories using pre-existing maps or bathymetric data, typically employing graph-search or sampling-based algorithms such as A*, D* and RRT. These paths serve as a coarse blueprint for {\em optimal} navigation. However, due to changing environmental conditions and sensor uncertainty, global plans frequently become outdated or infeasible mid-mission (\cite{ Guo2021, ishibashi2006}). This highlights the crucial need for a complementary planning approach to ensure continuous, safe and efficient vehicle operation. sensor-based path planning emerges as a critical component in this context. Operating on shorter timescales and smaller spatial horizons, sensor-based planners continuously update the UUVs immediate trajectory based on real-time sensor data (\cite{OKEREKE2023, Mingyue2022, RAFA2024}). This real-time capability is essential for dynamic and challenging uncertain environments, where obstacles, currents or other unpredictable factors necessitate instantaneous responses. sensor-based planning handles collision avoidance, trajectory refinement and safe maneuvering through cluttered or unknown areas, thus significantly enhancing mission safety and reliability.  

\begin{figure*}[ht]
    \centering
    \includegraphics[width=0.9\textwidth]{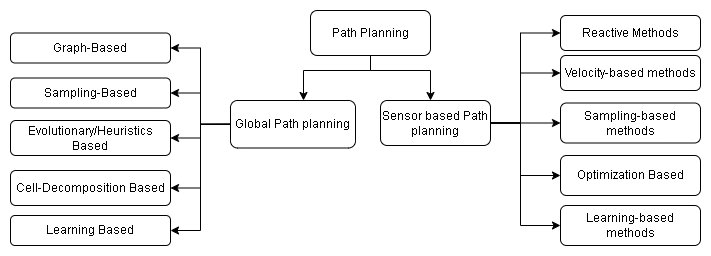}
    \caption{ Path Planning Techniques}
    \label{fig:path_plan}
\end{figure*}

The importance of sensor-based planning extends beyond simple obstacle avoidance. It is fundamental to the robustness of autonomous underwater navigation systems, providing adaptability in scenarios where global plans are compromised by changing environmental conditions or unforeseen obstacles. By continuously interpreting sensor inputs, sensor-based planning ensures that UUVs can rapidly adapt their trajectories to maintain operational safety, precision and efficiency, even in challenging underwater environments. This demands advanced sensor integration, robust filtering, accurate environmental modeling and reliable obstacle detection methodologies (\cite{QinYuan2023, Hernández2015, Liam2010}). Thus, sensor-based planning complements global planning and is indispensable for the practical deployment and success of autonomous underwater missions. Figure ~\ref{fig:path_plan} broadly reviews different global and sensor-based path planning techniques, each designed with a distinct objective to ensure safety and efficiency 
(\cite{Behnaz2022}).

Within the navigation stack of UUV systems, sensor-based planning and control frameworks can be categorized as either decoupled or coupled, depending on the level of integration between induced trajectory generation and actuation: 

\begin{itemize}

    \item In decoupled frameworks, the planning module operates independently from the control system. The planner generates trajectories based on expected ideal conditions, and the controller is subsequently responsible for effectively tracking these paths (\cite{Kennedy2007, Han2020, Volpi2018}). This approach provides significant advantages in terms of simplicity and modularity with optimal guarantees, facilitating independent design and tuning of planning and control algorithms. Although dynamic or uncertain conditions, such as strong currents or unforeseen obstacles, can sometimes pose challenges, these scenarios provide opportunities to enhance robustness by integrating adaptive strategies, frequent re-planning or advanced feedback mechanisms (\cite{Edward2022, Niankai2022}).


    \item The coupled frameworks combine the planning and control functions into one unified decision-making procedure. In this integrated approach, trajectory generation explicitly considers the vehicle's dynamic model, operational constraints and continuous sensor feedback (\cite{Yilmaz2008, Santhakumar2008}). Techniques like Model Predictive Control (MPC)(\cite{Mashhood2024}) and Invariant-set based Integrated Planning and Control (IPC) (\cite{Karthik2021, Karthik2024}) optimize trajectories and control inputs over a moving time window, ensuring adherence to constraints while simultaneously minimizing a predefined cost criterion. Such integrated methods typically offer computational efficiency and substantial improvements in adaptability, safety and robustness, particularly in uncertain environments.
\end{itemize}

Sensor-based planning and control is particularly indispensable in underwater robotics. Given the limitations of prior maps and the high variability of underwater terrain, UUVs must perceive and react to their surroundings autonomously (\cite{Ma2023}\cite{Petillot2021}\cite{Petillot1998}). This necessitates advanced techniques for processing sonar data, fusing information from multiple sensor modalities and building real-time representations of the environment. Popular representations include occupancy grids, 3-D point clouds, elevation maps and probabilistic models. All of which inform the local planner about safe and navigable regions (\cite{Zacchini2022, Yvan2001}). Yet, these capabilities come with their integration and computational challenges. Sonar data is often sparse, ambiguous and subject to multi-path reflections (\cite{Heshmati2020}). Synchronising and fusing information from sonar, DVL, IMU, and environmental sensors demands precise time stamping and real-time processing pipelines. Furthermore, onboard computing limitations restrict the complexity of algorithms that can be deployed on UUV platforms, requiring a balance between computational tractability and navigational robustness (\cite{Marin2018, Zhiqiang2022}).

Control approaches have also addressed the unique challenges posed by underwater environments (\cite{Heshmati2020, Zhouhua2019}). Most existing literature treats path planning and control as separate disciplines, often drawing from generic robotics (\cite{Lawrance2019, Chrpa2015, Pineda2018}). To explicitly address marine-specific constraints such as sensor degradation, buoyancy instability or communication bottlenecks, this survey intends to bridge that gap by offering a systematic sensor based planning and control strategies tailored for UUVs. We structure our discussion around the classification of navigation frameworks as decoupled or coupled, evaluating their performance, providing scenario-specific analysis of the coupled architecture, its complexity, adaptability and suitability for real-time operations. 

To support this analysis, we first present a conceptual foundation of underwater navigation methodologies. Figure~\ref{fig:path_plan} introduces a taxonomy of global and sensor-based path planning techniques, outlining their respective roles in enabling safe and efficient navigation. This taxonomy provides essential context for understanding how different planning strategies interact with control frameworks in sensor-driven underwater systems. 

Global path planners handle large-scale trajectory generation using map data and environment models. Common methods include graph-based approaches (e.g., A*), sampling-based planners like PRM and RRT, and heuristic techniques such as Genetic Algorithms \cite{chin2018}. Cell-decomposition divides areas into simpler regions for structured planning, while learning-based strategies use prior experience to generalize path generation. \\
In contrast, sensor-based planners operate reactively, updating movement in real time using sensor inputs. These include reactive behavior, velocity-space methods (e.g., DWA), sampling-based re-planners, optimization-based schemes like MPC, and machine learning approaches that adapt to environmental dynamics quickly \cite{Yvan2001}. By combining global foresight with local reactivity, these planning layers form a cohesive strategy essential for robust navigation in complex underwater environments. This taxonomy acts as the conceptual backbone for sensor-based autonomous control and sets the stage for the detailed discussions that follow in subsequent sections \cite{Hasan2024}.

The paper outlines the evolution of sensor-based underwater navigation, followed by a detailed description of the Underwater Autonomous Navigation Stack. It explores sensor-based planning and control frameworks, focusing on their integration and decision-making roles. A comprehensive analytical comparison is conducted between different coupled planner-controller architectures across complex underwater scenarios. Finally, the paper concludes by highlighting key insights and proposing future research directions to enhance the adaptability, robustness and performance of unmanned underwater vehicles (UUVs).

\begin{figure*}[ht]
\centering
\includegraphics[width=1.0\textwidth]{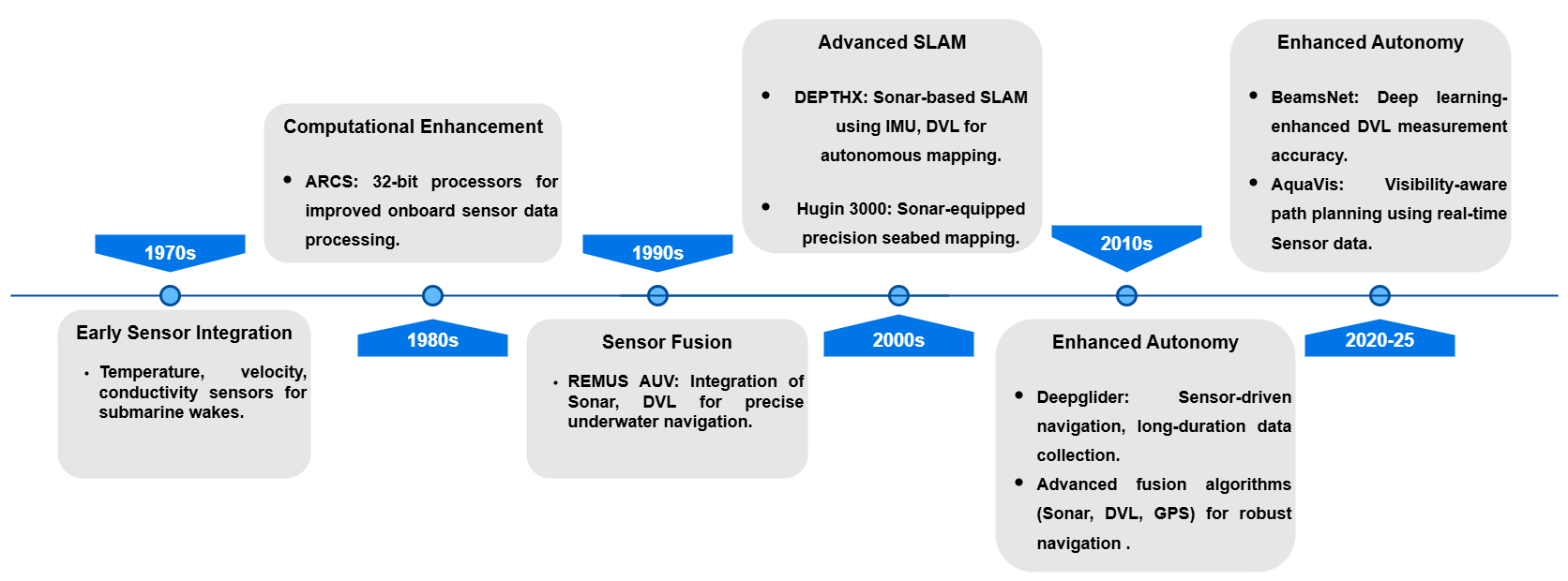}
\caption{ A chronological view of Major Technological Breakthroughs in Sensor-based Autonomous Underwater Navigation, its evolution from the 1970s to the 2025.}
\label{fig:Planning_timeline}
\end{figure*}

\section{Background}\label{sec2}

The field of underwater robotics has evolved substantially over the last six decades, driven largely by continuous innovations in sensor technology and intelligent control strategies. The ability of Unmanned Underwater Vehicles (UUVs) to navigate through unpredictable and intricate aquatic environments is fundamentally reliant on their perception capabilities and how dynamically they can respond to external stimuli. This section outlines the chronological evolution of sensor-driven navigation approaches, laying the groundwork for advanced planning and control architectures in modern UUVs (\cite{Paull2012, Yang2021}).  Figure \ref{fig:Planning_timeline} presents a timeline of technological breakthroughs that have shaped the current landscape of UUVs autonomy, from foundational control and acoustic localisation systems to modern AI-integrated planning architectures and endurance-focused vehicles like the Deepglider UUV \cite{zhang2020}.

The early developments in the \textbf{1970s} revolved around foundational sensing techniques. During this era, underwater systems utilized basic sensors to record environmental variables like temperature, flow velocity and conductivity—primarily aimed at detecting submarine wakes. While these measurements were limited in complexity, they laid the initial foundation for underwater situational awareness and enabled basic environmental profiling to inform movement decisions (\cite{Balestrieri2021}).

With the advent of more powerful processing capabilities in the \textbf{1980s}, the scope of underwater sensing expanded significantly. Notable among these developments was the integration of 32-bit processors in autonomous systems like ARCS, which enhanced onboard computation of sensor inputs. This shift enabled real-time data interpretation directly on the vehicle, improving response speed and supporting decentralized decision-making, a precursor to today’s embedded autonomous behaviors (\cite{ Valavanis1997}).

In the \textbf{1990s}, the concept of \textit{sensor fusion} came to the forefront. This decade witnessed the rise of platforms like REMUS UUVs (\cite{stokey2005}), which combined data from acoustic and inertial sensors, such as Doppler Velocity Logs (DVLs) and Sonar systems, to provide accurate navigation even in GNSS-denied settings. The fusion of different sensing modalities allowed for more reliable dead-reckoning and introduced the use of terrain-relative navigation—an approach that still underpins much of modern UUV localisation today (\cite{ Budiyono2009, Felemban2015}).

The progression into the \textbf{2000s}, brought a new wave of innovation through Simultaneous Localisation and Mapping (SLAM). Advanced platforms like DEPTHX (\cite{gary2008}) integrate multiple sensors—including IMUs, DVLs and multibeam sonar—to autonomously map unknown cave systems. Similarly, the HUGIN 3000 (\cite{vestgard2000}) system combined precise sonar capabilities with inertial measurements to deliver high-resolution seabed maps. These projects highlighted how a tight coupling of environmental perception with localisation could enable autonomous operations in previously unreachable regions (\cite{Antonelli2001}).

By the \textbf{2010s}, UUVs had evolved into systems capable of higher autonomy levels, guided by real-time sensor feedback. The Deepglider (\cite{osse2007}) platform exemplified this trend by navigating energy-efficient paths based on measured oceanographic data like salinity and temperature gradients. At the same time, sensor fusion algorithms advanced to handle complex terrain and varying depths by combining surface GNSS, inertial navigation and DVL data. This decade also saw the integration of machine learning in navigation, with tools like BeamsNet (\cite{OKEREKE2023}) refining velocity measurements and AquaVis (\cite{Mingyue2022}) enhancing route selection by factoring in water visibility and turbidity conditions .

Currently, in the \textbf{2020-25}, sensor-based autonomy has become more sophisticated and adaptive. The emphasis has shifted toward real-time, context-aware planning using techniques like deep reinforcement learning and predictive control algorithms. Platforms such as AquaVis (\cite{Mingyue2022}) showcase how incorporating live visual inputs can help UUVs dynamically adjust their path based on underwater visibility. Meanwhile, modern navigation frameworks simultaneously leverage data from sonar, DVL, vision systems and ocean current sensors, enabling UUVs to interpret their environments more holistically and perform reliably in previously uncharted or harsh underwater conditions (\cite{ QinYuan2023, Orłowski2022}).

\begin{table*}[ht]
\centering
\caption{Widely used Planning and Control Methods in UUVs}
\renewcommand{\arraystretch}{1.25}
\resizebox{\textwidth}{!}{
\begin{tabular}{@{}p{3cm}p{5.5cm}p{5.5cm}p{1cm}p{2cm}@{}}
\toprule
\textbf{Planner/Control Method} & \textbf{Significance} & \textbf{Limitations} & \textbf{Year} & \textbf{Referances} \\
\midrule

Dubins Path + PID & 
Provides a lightweight and easy-to-implement solution for fixed-curvature path tracking in known environments. Widely used in structured scenarios such as pipeline inspection or area coverage. & 
Incapable of responding to sudden environmental changes or dynamic obstacles. Performance drops significantly in turbulent or unknown environments. Due, to limited disturbance rejection capability & 
2007 & \cite{Pêtrès2007} \\


Deep RL-based Planning & 
Enables learning of control policies directly from environmental interactions, allowing adaptability in highly dynamic and uncertain underwater scenarios. & 
Demands large-scale data and extensive training, which may be infeasible in remote deployments. Computational load hinders real-time use on embedded systems. & 
2022 & \cite{Behnaz2022} \\

CBF + MPC & 
Combines constraint enforcement via Control Barrier Function (CBF) with predictive optimisation via Model Predictive Control (MPC), enabling real-time decision-making that ensures safety in proximity to obstacles and variable currents. & 
Requires high computational resources, limiting deployment on low-power embedded systems. Real-time performance depends on solver efficiency. & 
2020 & \cite{QinYuan2023} \\

Learning-based Terrain-Aided Navigation & 
Leverages terrain features and deep models to improve localisation in GNSS-denied environments. Suitable for long-term missions in topographically rich underwater regions. & 
Relies on accurate prior terrain maps or effective feature extraction. Performance questionable in flat or textureless seafloor regions. & 
2023 & \cite{Heshmati2020} \\

Integrated Planning and Controls (IPC) & 
Facilitates the joint design of planning and control, minimising trajectory-tracking errors through an integrated model. Enhances robustness in multi-disturbance scenarios. & 
 Optimal performance is questionable. & 
2024 & \cite{Karthik2021, Karthik2024} \\

\bottomrule
\end{tabular}
}
\label{tab:uuv_planning_controls}
\end{table*}

As sensors have become more capable, the need for equally intelligent planning and control strategies has become evident. These strategies govern how UUVs interpret sensor data and respond through motion and path correction. Table \ref{tab:uuv_planning_controls} highlights some of the most widely used planning and control methods in contemporary underwater vehicles and renders the challenges involved in developing an integrated framework that can handle various underwater scenarios.

A recent innovation involves learning-based terrain-aided navigation, which leverages terrain elevation maps and learned models to support localisation in feature-rich seabeds. These methods have shown high promise in long-term deployments for scientific exploration and military reconnaissance \cite{Mingyue2022}. Their reliance on prior data and feature extraction, however, poses challenges in sparse or textureless environments. In parallel, Integrated Planning and Control (IPC) (\cite{Karthik2021, Karthik2024}) methods have recently been developed in which IPC frameworks unify induced trajectory generation and control execution within a single loop, reducing reaction delays and optimising performance against predefined objectives. These techniques outperform traditional modular designs, especially when dealing with variable mission goals and unforeseen events. However, they have not yet established reliable  results. This visual summary in fig \ref{fig:Planning_timeline} complements the comparative table (Table \ref{tab:uuv_planning_controls}), which captures prominent planner-control strategies, their practical impact and inherent limitations.

Together, these milestones highlight the transition from decoupled and rule-based modules to robust, integrated architectures capable of real-time decision-making in uncertain marine environments (\cite{Orłowski2022}). As underwater operations becomes more complex and autonomous missions become the norm, the focus continues to shift toward hybridised systems that offer safety, resilience and adaptability hallmarks of the next generation of underwater robotics.

The convergence of these historical sensor innovations and modern control strategies has enabled the formulation of modular, layered architectures that form the backbone of underwater autonomy. These architectures, often termed as \textit{navigation stacks}, include dedicated subsystems for perception, localisation, mapping, planning and control (\cite{QinYuan2023, Mingyue2022, Orłowski2022, Liam2010}). Each layer relies on sensor feedback and algorithmic decision-making to fulfil specific roles in the mission. This historical perspective shows how sensor technology has driven the evolution of underwater navigation, from simple environmental monitoring to intelligent autonomy powered by real-time, multi-modal sensor fusion. With the advancement of integrated control frameworks, modern UUVs are no longer reactive tools but intelligent agents capable of autonomous exploration, inspection and intervention. The next section presents the Autonomous Underwater Navigation Stack, detailing its functional components and illustrating how sensor data and planning methods work together to ensure reliable, real-time navigation in dynamic and uncertain underwater environments.

  \begin{table*}[ht]
    \centering
    \caption{Comparison of Underwater Communication Technologies }
    \label{tab:underwater_comm}
    \renewcommand{\arraystretch}{1.3}
    \begin{tabular}{@{}p{3.2cm}p{3cm}p{5cm}p{5cm}@{}}
    \toprule
    \textbf{Technology} & \textbf{Direction} & \textbf{Advantages} & \textbf{Disadvantages} \\
    \midrule
    Optical & Local directional & 
    Very high data rates (up to hundreds of Mbps), Low-cost implementation. &
    Highly sensitive to turbidity and particle presence, Requires precise alignment, Very limited range. \\
    
    Radio & Omni-directional & 
    Can penetrate ice and turbid water, Immune to multipath effects, Offers high bandwidth (from several kbps to Mbps). &
    Easily disrupted by electromagnetic interference, Significantly limited range underwater. \\
    
    Electro-communication & Omni-directional & 
    Compact in size, Energy efficient, Moderate bandwidth, Cost-effective. &
    Limited transmission distance. \\
    
    Acoustic & Omni-directional & 
    Long operational range (up to 20 km), Proven and mature technology. &
    Lower bandwidth, Subject to latency and ambient noise. \\
    \bottomrule
    \end{tabular}
    \end{table*}

\section{Autonomous Underwater Navigation}\label{sec3}

Autonomous navigation in Unmanned Underwater Vehicles (UUVs) is a cornerstone for achieving reliable and efficient subsea operations without human intervention. These vehicles are instrumental across various applications including seabed mapping, underwater archaeological surveys, environmental monitoring, pipeline inspection and defense-related reconnaissance missions (\cite{Thale2020, Bijjahalli2020, Song2003}). The advancement of UUVs into truly autonomous systems relies heavily on sophisticated subsystems that collectively allow the robot to perceive, communicate, plan and act without continuous human oversight. This progression toward full autonomy is governed by a navigation stack, a layered architecture encompassing perception, communication, localization, path planning and control. Each of these layers communicates with and informs the others, forming a feedback-driven loop that enables the UUVs to function reliably in complex underwater environments (\cite{Eickstedt2010, Qin2022}).

      \subsection{Underwater Communication Technologies}

    Effective communication is a critical enabler in the operation of Unmanned underwater vehicles (UUVs), particularly for tasks involving remote supervision, cooperative behaviors, mission updates and data transmission. The challenging nature of the underwater medium, marked by high signal attenuation, variable propagation speeds and multi-path effects, places significant limitations on conventional communication methods (\cite{Yang2021}). Therefore, a good understanding of underwater communication technologies is essential for selecting and deploying the most suitable approach based on mission objectives, environmental conditions and system architecture.

    The table ~\ref{tab:underwater_comm} compares four primary underwater communication methods: Optical, Radio, Electro-communication and Acoustic. Each technology addresses a specific operational need and offers a distinct trade-off between range, bandwidth, directionality and environmental robustness.

    \textbf{Optical Communication} links deliver extremely high bandwidth—often reaching hundreds of megabits per second—and are ideal for applications requiring the transfer of large datasets, such as imagery, video or multi-beam sonar scans. They are also cost-effective and relatively easy to implement in clear water conditions (\cite{Immas2022}). However, optical systems are limited by turbidity, alignment sensitivity and range, often capped at a few meters. These limitations restrict their use to short-range, high-speed data bursts, such as when UUVs return to a docking station or communicate with nearby swarm units.
    
    \textbf{Radio Communication}, though effective in air, face severe attenuation in water. Only low-frequency RF signals can propagate, resulting in limited range but moderate data rates. Their key advantage is their ability to penetrate ice and turbid water—a critical feature in polar exploration or post-disaster search operations. Radio communication is also resilient to multi-path interference, which benefits signal integrity. Still, due to electromagnetic noise and limited underwater range, its application is often confined to shallow water or surface-near operations (\cite{keerthi2020}).
    
    \textbf{Electro-communication} are inspired by electric fish, electro-communication offers low-power, compact data transmission suitable for short-range swarm coordination or proximity sensing. Its simplicity and energy efficiency make it appealing for small-scale deployments or secondary backup channels (\cite{wang2015}). Its main limitation is its extremely limited range and high sensitivity to the environment’s electrical conductivity. Its use remains limited to very specific UUV applications.
    
    \textbf{Acoustic Communication} systems dominate the underwater communication space due to their ability to transmit signals over distances of up to 20 km. They offer omnidirectional coverage, mature technological infrastructure and reliable operation across various mission profiles (\cite{Shields2023}). The trade-off lies in low bandwidth, high latency and vulnerability to ambient noise. However, these issues are often mitigated through encoding schemes, error correction protocols and smart routing algorithms. Acoustic systems are indispensable for long-range and deep-sea UUV operations.

\begin{table*}[ht]
    \centering
    \caption{Comparison of UUVs localisation Techniques  }
    \label{tab:auv_nav_methods}
    \renewcommand{\arraystretch}{1.4}
    \begin{tabular}{@{}p{3.75cm}p{5.5cm}p{7.5cm}@{}}
    \toprule
    \textbf{Localization Methods} & \textbf{Advantages} & \textbf{Disadvantages} \\
    \midrule
    \textbf{Dead Reckoning} & 
    Autonomous operation with consistent tracking and decent short-term precision & 
    Prone to cumulative drift over time, leading to large positioning errors in extended missions \\
    \textbf{Signal-Based localization Techniques} (e.g., Sonar, LBL) & 
    High localization accuracy and immunity to drift enable precise positioning without cumulative error & 
    Susceptible to signal degradation, interference and obstruction; requires infrastructure and has limited range \\
    \textbf{Map-Matching Navigation} & 
    Drift-free operation and independence from external signals & 
    Heavily reliant on the availability of accurate and recent maps; high processing demand; performance varies with environmental complexity \\
    \bottomrule
    \end{tabular}
    \end{table*}

     \subsection{Underwater Localization Methods}

    The underwater environment presents a complex and uncertain setting that necessitates the use of highly specialized localization strategies for UUVs \cite{Paull2014}. Unlike aerial or ground robotics, where satellite-based localization is standard, underwater navigation relies on sensor-based inference, acoustic positioning systems and terrain referencing techniques to estimate and update vehicle position. In this context, the analysis of UUV localization technique plays a foundational role in both the design and operational performance of autonomous underwater systems. 

    The table ~\ref{tab:auv_nav_methods} classifies three primary navigation methods; Dead Reckoning, Signal-Based Navigation and Map-Matching Navigation; each offering a distinct balance of performance, resource dependency and operational constraints (\cite{Tan2011, Thale2020}).

    \textbf{Dead reckoning} is one of the oldest and simplest navigation strategies, estimates current position based on velocity and heading data from an initial known position. This technique benefits from being entirely self-contained, with no reliance on external infrastructure or environmental features. It provides good continuity and short-term precision, making it ideal for missions with constrained durations or predictable conditions. However, its major drawback lies in error accumulation (\cite{ Danovaro2014}). As position estimates are continuously updated based on internal measurements, any small errors in sensor readings, particularly from inertial systems, lead to growing positional drift. Over extended missions, this cumulative drift can severely impair mission outcomes, making dead reckoning alone unsuitable for long-range deployments. Nevertheless, it remains a foundational component in fused navigation frameworks, particularly where external references are temporarily unavailable.
    
    \textbf{Signal-based navigation} methods, including Long Baseline (LBL), Short Baseline (SBL), Ultra-Short Baseline (USBL) and sonar-based positioning, use acoustic signals to triangulate the UUV position relative to external beacons or reflective surfaces. This approach yields high localization accuracy and is immune to accumulated error, making it a popular choice for precision-critical tasks such as infrastructure inspection, docking or geo-spatial surveys. Despite its advantages, signal-based navigation suffers from dependency on external infrastructure and environmental susceptibility (\cite{Hidalgo2015}). Acoustic signals degrade over distance and are impacted by water temperature, salinity and noise. Systems like LBL also require pre-deployed transponders, adding logistical overhead. Furthermore, signal-based systems are vulnerable to jamming or multi-path distortion, especially in environments with high acoustic clutter. Nonetheless, the deterministic precision and drift-free performance of signal-based methods make them invaluable for tasks where safety, accuracy or repeatability are non-negotiable (\cite{Panda2020}).
    
    \textbf{Map-matching} strategies use environmental features captured during mission execution (e.g., sonar images, depth profiles) and align them with previously stored maps to determine the vehicle’s location. This method offers drift-free operation and functions well even in the absence of external signals like GNSS or acoustic beacons. It is particularly useful in structured or semi-structured areas with distinguishable terrain patterns, such as underwater pipelines or canyon-like seafloor. The key challenge with map-matching lies in its dependency on high-quality, up-to-date maps and significant computational requirements (\cite{Koslow2007, Song2003}). If the operating environment has changed since the reference map was created, localization accuracy can deteriorate. Additionally, in regions with little to no distinctive features—such as flat seabeds, the matching algorithm may struggle to converge. Despite this, map-matching is a critical tool in the underwater localization, often used in conjunction with SLAM or signal-based methods for robustness.

    The reviewed localization strategies and communication technologies collectively form the operational basis for situational awareness of the autonomous underwater vehicles. Tables  ~\ref{tab:underwater_comm} and ~\ref{tab:auv_nav_methods}  highlight the critical role of multi-modal sensing and adaptive communication in addressing the diverse demands of underwater missions. Effective system selection requires a thorough assessment of mission goals, environmental challenges, and platform capabilities—often leading to the adoption of hybrid architectures that leverage complementary strengths for resilient, long-duration UUV operations in complex marine environments \cite{williams2000, mallios2014}. The situational awareness derived from localization and perception systems forms the foundation for subsequent sensor-based path planning and control processes. 

    \subsection{Autonomous Navigation Stack}

\begin{figure}[ht]
    \centering
    \includegraphics[width=0.5\textwidth]{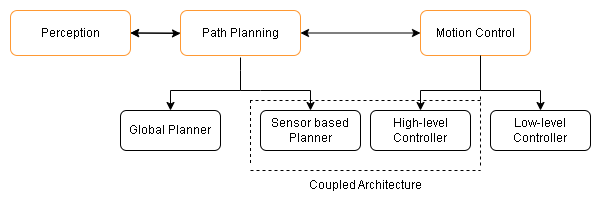}
    \caption{ Elements of UUV navigation stack}
    \label{fig:navigation_stack}
    \end{figure}

    The underwater autonomous navigation stack is structured as a hierarchy of interdependent modules, each dedicated to a core functionality. Together, they empower the UUV to carry out long-duration missions while continuously adapting to new environmental stimuli and sensor readings. Figure~\ref{fig:navigation_stack} illustrates the modular breakdown of the stack, emphasizing the tight coupling between perception, path planning and control \cite{Leonard2013}.

    \textbf{Perception} serves as the foundational layer. It provides the sensory inputs needed to understand the UUVs environment. The perception subsystem is primarily composed of sonar-based systems, Doppler Velocity Logs (DVLs) and sometimes optical cameras and acoustic modems. Sonar technologies, such as multibeam echosounders and side-scan sonar, enable terrain mapping and obstacle detection. DVLs and Inertial Measurement Units (IMUs) contribute to dead reckoning and relative localisation. The data from these sensors is filtered and interpreted using signal processing and feature extraction techniques to identify relevant features such as seafloor contours, obstacles and navigational waypoints (\cite{Chang2022}).
    
    \textbf{Path Planning} is the next layer and acts upon the environmental model constructed by the perception system \cite{Yao2019, Yilmaz2008}). It is generally divided into two major subdomains:
    \begin{itemize}    
    
        \item \textbf{Global Path Planning:} Operating over large mission timescales, this module uses prior bathymetric maps or mission-specific goals to compute a collision-free trajectory from the start location to the target. Common algorithms include A*, D*, Probabilistic Roadmaps (PRM) and Rapidly-Exploring Random Trees (RRT). These plans serve as high-level navigational guides (\cite{Pailhas2007, Meng2025}).    
        
        \item \textbf{Sensor-based Path Planning:} This module handles near-term, real-time decision-making using current sensor inputs. It deals with dynamic obstacle avoidance, terrain adaptation and short-horizon trajectory refinement. Algorithms such as the Dynamic Window Approach (DWA), Curvature Velocity Method (CVM) and Artificial Potential Fields (APF) are commonly employed (\cite{Mingyue2022, Liam2010}).
        
    \end{itemize}
    
    \textbf{Control} is the next layer of the navigation stack and is responsible for ensuring that the vehicle follows the desired path. \cite{Budiyono2009}
    There are two major paradigms in control design for UUVs:
    
    \begin{itemize}    
        \item \textbf{High-level Control:} This component of the navigation architecture is tasked with generating the appropriate velocity commands required to guide the vehicle toward an intermediate goal point provided by the planner. It serves as the bridge between high-level planning and low-level actuation, translating spatial goals into executable control signals such as surge, pitch and yaw rates. By continuously evaluating the current state of the vehicle and the desired target, this module ensures that the vehicle follows a smooth and feasible trajectory, adhering to dynamic and operational constraints while progressing reliably toward the next waypoint \cite{RAFA2024}.    
        \item \textbf{Low-level Control:} This module within the architecture is responsible for locally refining and regulating the high-level commands generated by the planner. Its primary function is to minimise tracking errors in real time, enabling the system to make continuous adjustments to the vehicle's motion. By doing so, it helps reduce deviation from the intended path and ensures more accurate and stable navigation. This local control layer acts as a corrective mechanism, responding dynamically to environmental disturbances, model uncertainties or sudden changes in trajectory, thereby enhancing the overall reliability and precision in reaching the designated goal \cite{Zhiqiang2022}.
        
    \end{itemize}
Fig ~\ref{fig:navigation_stack} renders the schematic UUV Navigation Stack in the context of integration possibility explored in the literature. The coupled architecture typically integrates sensor-based planning and high-level control for efficient reactive operations.
The navigation stack is dynamic, allowing continuous feedback and updates. For example, sonar-based perception may detect an obstacle that triggers a re-planning process at the local level while maintaining alignment with global objectives. The coupling of local planners with reactive controllers facilitates timely responses to disturbances such as ocean currents or unanticipated terrain. It is clear that advanced techniques such as Terrain-Aided Navigation (TAN) and real-time SLAM (Simultaneous Localization and Mapping) are still embedded within the stack, enabling drift-free localization and navigation in GNSS-denied environments. Data from sonar or DVL sensors is used to construct or update occupancy grids or point cloud maps, informing the decision-making layer about safe and unsafe zones (\cite{ Raju2020, Koslow2007, Bijjahalli2020, Mingyue2022}).

\begin{figure}[ht]
    \centering
    \includegraphics[width=0.5\textwidth]{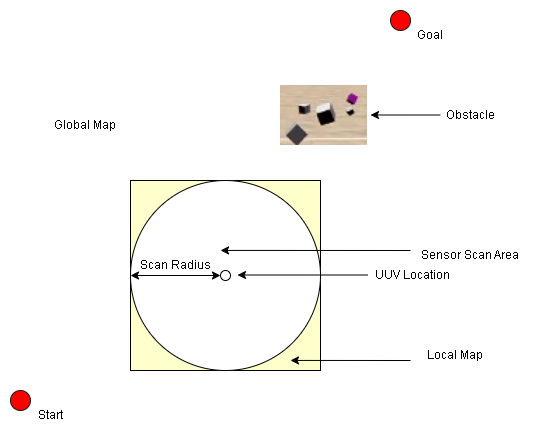}
    \caption{Illustrative model outlining the typical UUV operational setting in an underwater setting}
    \label{fig:uuv_env}
    \end{figure}

    The accompanying figure ~\ref{fig:uuv_env} conceptually illustrates how an UUV operates in an underwater environment within a layered navigation structure comprising a global map and a sensor-based, local scanning map. In this representation, the UUV begins its mission at a designated start location and aims to reach a predefined goal point, navigating around static or dynamic obstacles on its way (\cite{Hidalgo2015}).

   Figure~\ref{fig:uuv_env} depicts a sensor-based map overlaid on the larger global map. The global map represents the mission-level environment, often pre-constructed from bathymetric data or previously acquired maps. It outlines macro-level features such as terrain contours, static infrastructure and known obstacles. However, it cannot reflect real-time environmental variations like drifting objects, marine life or dynamic sediment plumes (\cite{Mingyue2022, Zacchini2022}).

    This limitation is addressed by incorporating a sensor-based, local scanning area, typically represented as a circular sensing zone centred around the UUV in operation. This area corresponds to the effective range of the onboard sensors, such as sonar, LIDAR (if applicable) or Doppler Velocity Logs (DVL). The radius of this circle defines the reactive horizon for the vehicle and is critical for tasks such as immediate obstacle avoidance, short-term trajectory adjustment and terrain-following behaviour (\cite{Yvan2001, Marin2018}).
    In practice, the UUVs continuously updates this sensor-based map using live sensor data, ensuring that its reactive path planner has the most up-to-date situational awareness. This dynamic interplay between the static global map and the responsive sensor-based map is what allows the UUVs to balance strategic mission planning with tactical reactivity (\cite{Zhiqiang2022}). It also supports the use of hybrid planners, which combine global path strategies with local reactive controllers. The feedback loop between sensing, planning and control enables real-time trajectory adaptation in complex environments.
    
    The reasoning for this architectural design stems from the fact that underwater conditions are inherently uncertain. UUVs cannot rely solely on prior information, and continuous perception is essential. By segmenting perception and planning into global and sensor-based domains, the system can operate efficiently over long missions while remaining agile in the face of local perturbations (\cite{OKEREKE2023}).

\section{Sensor-based Planning and Control}\label{sec4}

Sensor-based planning and control have become a foundational aspect of enabling true autonomy in Unmanned Underwater Vehicles (UUVs). In the absence of consistent global positioning and reliable communication, UUVs must depend entirely on onboard sensors for environmental perception and navigation. This paradigm centers around real-time data acquisition from sources such as multi-beam sonars, Doppler Velocity Logs (DVL), inertial measurement units (IMUs), optical cameras and pressure sensors, which together provide a continuous stream of feedback essential for path generation and control law execution.

In sensor-based planning, UUVs dynamically compute motion trajectories by processing environmental data to evaluate obstacle-free regions, detect terrain features and adapt to changing operational constraints \cite{Antonelli2001}. These systems replace static, pre-computed paths with real-time adaptability, which is critical for operations in unknown or cluttered underwater terrains. Algorithms like Rapidly-exploring Random Trees (RRT) \cite{Zacchini2022} and their sensor-augmented variants are employed to reactively update paths, ensuring mission continuity in the face of disturbances or new obstacle detections. On the control side, sensor data feeds directly into decision-making loops to refine trajectory tracking, maintain vehicle stability and ensure adherence to physical constraints. Techniques such as Model Predictive Control (MPC) (\cite{Zhongrui2025}) and Integrated Planning and Control (IPC) \cite{ Karthik2024} leverage real-time measurements to correct deviations and enforce safety boundaries. These frameworks provide the flexibility to adapt motion commands in response to current flow, terrain gradients or unexpected obstacles—all detected through continuous sensing.

As sensor technologies have matured, their integration with adaptive control policies has enabled UUVs to autonomously operate in high-risk and GNSS-denied environments. This section explores key sensor-based planning and control methodologies, highlighting how real-time sensor interpretation governs both the generation of feasible paths and the precise execution of motion commands, ultimately ensuring safety, reliability and mission success in complex underwater domains.

    \subsection{Coupled and Decoupled Planning and Control Architectures} 

    A sensor-based navigation in underwater robotics hinges on the integration of sensing, planning and control systems. Two fundamental paradigms govern this interaction: \textbf{decoupled} and \textbf{coupled} architectures. In a decoupled framework, the system is organised into two distinct layers—path planning and control. The planner first generates an optimal trajectory considering environmental maps and mission goals, assuming ideal execution (\cite{Kennedy2007, Morgan2022}). The controller then attempts to follow this predefined path, typically using control strategies like PID and Lyapunov band control. These modules operate independently, simplifying the design and offering modularity for developers.

    In contrast, coupled architectures blend planning and control into a unified decision-making system. Here, the robot continuously adjusts its trajectory in real time by incorporating sensory feedback, environmental dynamics and vehicle constraints (\cite{Volpi2018, Jalal2021, Junzhi2021}). This tightly integrated framework (\cite{Guo2020}) allows the robot to anticipate future events and optimize control actions using advanced strategies such as Model Predictive Control (MPC) or Integrated Planning and Control (IPC).
    

    Understanding these architectures is essential not only for selecting the appropriate navigation stack but also for designing scalable systems capable of evolving into hybrid or adaptive frameworks. In the following discussion, we study architecture in depth to distinguish the functionality between the coupled and decoupled architectures, including their operational models, and environmental interactions. We then propose scenario-based study to evaluate performance of coupled architecture.

        \subsubsection{Decoupled Architecture}

\begin{figure}[ht]
            \centering
            \includegraphics[width=0.3\textwidth]{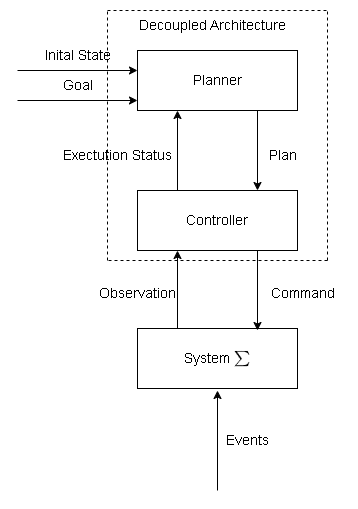}
            \caption{Illustrative model outlining the typical Decoupled Architecture}
            \label{fig:decop_arch}
            \end{figure}

         In Figure~\ref{fig:decop_arch}, the decoupled architecture separates planning and control into distinct sequential modules. The planner computes a trajectory or strategy based solely on the initial state and goal, without direct integration of environmental changes. This plan is then passed to the controller, which independently tries to track the trajectory and issues commands to the system (\cite{Kennedy2007}).

            Although observations and execution status are still relayed back to the planner, they are generally not used for real-time plan alteration. Instead, the control system attempts to mitigate disturbances through feedback loops, but the re-plan possibility stays with the planner  in dynamic conditions (\cite{Fenucci2024, Aguado2021}). The clear modularity allows for simpler development, optimal plans and testing, but can result in reduced robustness when operating in unstructured underwater environments where faster reactive planning is a requirement.

        \subsubsection{Coupled Architecture}

\begin{figure}[ht]
            \centering
            \includegraphics[width=0.5\textwidth]{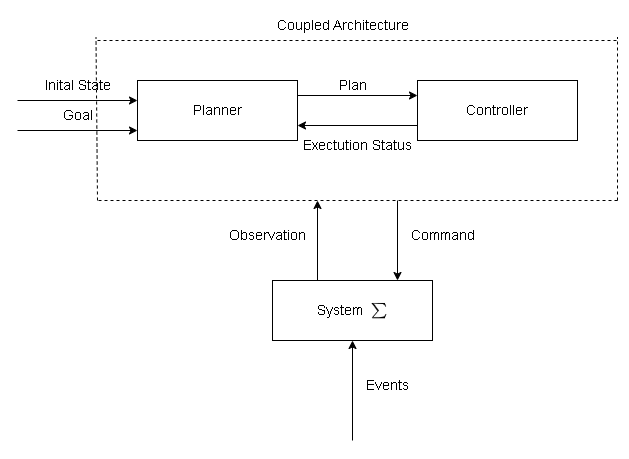}
            \caption{Illustrative model outlining the typical Coupled Architecture}
            \label{fig:cop_arch}
            \end{figure}

            In the coupled architecture, both the planner and controller are embedded within a tightly integrated loop. As shown in the Figure ~\ref{fig:cop_arch}, the initial state and goal are fed into the planner. Simultaneously, it receives observations from the real-world System \(\sum\)\, which includes sensors, environment, and vehicle states. The planner dynamically generates plans based on this feedback and directly interacts with the controller, providing updated paths or commands (\cite{ Zhongrui2025, Santhakumar2008}).

            Importantly, the execution status—how the plan is unfolding is returned to the planner, allowing real-time re-planning or refinement. This two-way feedback loop makes the coupled system adaptive and responsive to sudden disturbances, such as changing ocean currents or unanticipated obstacles. Events in the environment are directly perceived through the system, triggering updates in planning and execution (\cite{Pairet2021}). With this foundational understanding, the next section will suggests a scenario-specific study to analyze performance of various coupled planners and controllers. The evaluation will quantitatively measure the performance of each approach using key metrics such as path tracking error, goal-reaching success and time to reach, and resilience to disturbances. Additionally, it will qualitatively assess their suitability across a range of underwater scenarios. 

            
\subsection{Scenario-specific Study on Coupled Methods}\label{sec5}

The evaluation of autonomous navigation systems for underwater vehicles must extend beyond theoretical frameworks to encompass practical performance in realistic and diverse scenarios.  In this section, we perform a comparative study of different coupled planning-control architectures, specifically under environmental constraints characterized by cluttered terrains, dynamic obstacles and geometrically constrained regions such as infinitely long corridors or infinitely wide barriers.  These cases are chosen to simulate high complexity operational zones where traditional planning and control strategies are often stressed. To assess each approach systematically, we utilize core evaluation metrics including acceleration profiles, path curvature variations and total time to completion. These parameters directly impact the efficiency, smoothness and reactivity of the vehicle's trajectory and are essential for understanding the trade-offs involved in architecture design.

This comparative study evaluates planner-control architectures in sensor-based UUV navigation, focusing on the trade-offs between different controller design while utilizing the same integrated planner (IPC) (\cite{Karthik2024}) across all scenarios to ensure consistency in planning strategy. Assesses well each model integrates real-time sensor feedback to adapt to changing conditions. The goal is to identify scenarios in which a particular combination offers superior performance over other. These insights will inform a recommendation matrix to guide architecture selection based on mission demands and computational constraints.\\
Scenarios chosen for comparison are described below : \\

\begin{figure}[ht]
            \centering
            \includegraphics[width=0.45\textwidth]{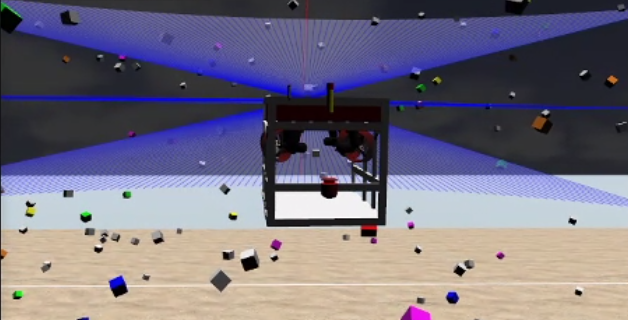}
            \caption{Scenario 1 : Underwater Environment with non-uniformly distributed unknown static obstacles }
            \label{fig:clu}
\end{figure}

\textbf{Scenario 1} : This scenario (Figure ~\ref{fig:clu}) represents an underwater environment populated with  unknown static obstacles that are non-uniformly scattered across the operational space. The mission objective is to navigate the vehicle toward a goal positioned ahead at the same depth level. Such a setting simulates realistic conditions commonly encountered in subsea operations, where the presence of natural formations, marine debris, coral structures, or stationary aquatic life creates a cluttered and complex terrain. The scenario is designed to evaluate the system’s ability to perform reliable path planning and control in environments where obstacle distribution is unpredictable and densely packed, necessitating precise maneuvering and high responsiveness from the navigation stack.

\begin{figure}[ht]
            \centering
            \includegraphics[width=0.45\textwidth]{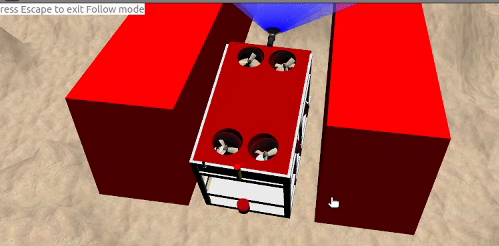}
            \caption{Scenario 2 : Underwater Environment with a static narrow passage }
            \label{fig:narrow}
\end{figure}

\textbf{Scenario 2} : This scenario (Figure~\ref{fig:narrow}) depicts an underwater environment where the vehicle encounters a narrow passage directly in its path. The passage is just wide enough to allow the vehicle to pass through safely, requiring precise control and maneuvering. This setup is intended to emulate real-world operational challenges such as navigating through submerged tunnels, between closely spaced underwater structures, or alongside pipelines and seabed installations. The scenario is particularly useful for testing the vehicle's ability to accurately align itself with constrained trajectories, maintain stability in tight spaces, and make controlled adjustments to avoid collisions. It serves as a critical benchmark for evaluating the effectiveness of both sensor-based planning and reactive control mechanisms in confined underwater environments.

\begin{figure}[ht]
            \centering
            \includegraphics[width=0.45\textwidth]{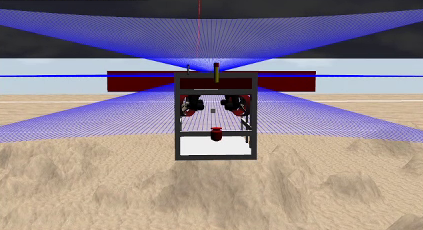}
            \caption{Scenario 3 : Underwater Environment with a infinitely wide static column ahead of the vehicle.}
            \label{fig:wide}
\end{figure}

\textbf{Scenario 3} : This scenario, as illustrated in Figure~\ref{fig:wide}, shows an underwater environment in which the vehicle is faced with a wide horizontal barrier obstructing its forward motion. This structure spans across the vehicle’s intended path, effectively acting as an unbounded barrier. Such a configuration simulates real-world situations where the vehicle may encounter expansive underwater formations—such as submerged cliffs, ship hulls or geological walls—that block direct access to the goal. To successfully navigate this scenario, the vehicle must dynamically adjust its planned trajectory, identify feasible vertical detours and demonstrate reactive capabilities to circumvent the obstruction. The objective remains to reach a goal located in front direction at the same depth, making this a critical test case for evaluating adaptive path planning and control algorithms in the presence of large-scale environmental constraints.
\begin{table*}[t]
    \centering
    \caption{Performance Metrics Across Navigation Scenarios}
    \label{tab:performance_metrics}
    \renewcommand{\arraystretch}{1.5}
    \footnotesize

    \resizebox{\textwidth}{!}{%
    \begin{tabular}{p{3.0cm} p{3.6cm} | r r | r r | r r }
        \hline
        \multirow{2}{*}{\textbf{Scenarios}} & \multirow{2}{*}{\textbf{Controller}}  
        & \multicolumn{6}{c}{\textbf{Planner and Controller Metrics}} \\
        \cline{3-8}
        & & \textbf{Tracked Path Length(m)} & \textbf{Time Taken(s)} 
        & \textbf{Avg Accel.} & \textbf{Max Accel.} & \textbf{Avg Curv.} & \textbf{Max Curv.} \\
        \hline

        \multirow{3}{*}{Cluttered Obstacles} 
        & MPC & 155.82 & 180.82 & 0.204 & 3.75 & {\fontsize{9.5pt}{12.5pt}\selectfont \textbf{0.027}}  & {\fontsize{9.5pt}{12.5pt}\selectfont \textbf{0.873}}  \\
        & PID & 171.12 & 270.46 & 0.226 & 3.46 & 0.069 & 3.800 \\
        & Invariant-set & 140.04 & {\fontsize{9.5pt}{12.5pt}\selectfont \textbf{162.74}} & {\fontsize{9.5pt}{12.5pt}\selectfont\textbf{0.131} } & {\fontsize{9.5pt}{12.5pt}\selectfont \textbf{3.00}}& 0.055 & 2.920 \\[2.0ex]

        \multirow{3}{*}{Narrow Passage} 
        & MPC & 156.67 & {\fontsize{9.5pt}{12.5pt}\selectfont \textbf{189.87}}  & {\fontsize{9.5pt}{12.5pt}\selectfont \textbf{0.053}}  & {\fontsize{9.5pt}{12.5pt}\selectfont \textbf{0.846}}  & 0.080 & {\fontsize{9.5pt}{12.5pt}\selectfont \textbf{1.970}}  \\
        & PID & 172.77 & 276.49 & 0.105 & 3.90 &  {\fontsize{9.5pt}{12.5pt}\selectfont \textbf{0.030}} & 2.980 \\
        & Invariant-set & 166.56 & 294.70 & 0.097 & 3.70 & 0.038 & 2.980 \\[2.0ex]

        \multirow{3}{*}{Wide Barrier} 
        & MPC & 155.72 &{\fontsize{9.5pt}{12.5pt}\selectfont  \textbf{180.71}} & 0.204 & {\fontsize{9.5pt}{12.5pt}\selectfont  \textbf{3.64}} & 0.026 &  {\fontsize{9.5pt}{12.5pt}\selectfont \textbf{0.875} }\\
        & PID & 148.09 & 238.35 & {\fontsize{9.5pt}{12.5pt}\selectfont \textbf{0.071} }  & 3.98 & {\fontsize{9.5pt}{12.5pt}\selectfont\textbf{0.017} }  & 0.990 \\
        & Invariant-set & 148.03  & 263.70 & 0.082 & 3.802 & 0.042 & 3.770 \\[2.0ex]

        \multirow{3}{*}{Long Barrier} 
        & MPC & 156.18 & {\fontsize{9.5pt}{12.5pt}\selectfont  \textbf{180.89} } & 0.179 & 3.95 & 0.026 & 0.860 \\
        & PID & 151.54 & 247.60 &  {\fontsize{9.5pt}{12.5pt}\selectfont  \textbf{0.058}} & {\fontsize{9.5pt}{12.5pt}\selectfont \textbf{2.93} } & {\fontsize{9.5pt}{12.5pt}\selectfont \textbf{0.017} }  & {\fontsize{9.5pt}{12.5pt}\selectfont  \textbf{0.780} } \\
        & Invariant-set & 149.78 & 266.02 & 0.368 & 3.36 & 0.018 & 0.830 \\[2.0ex]

        \multirow{3}{*}{Dynamic Obstacles} 
        & MPC & 156.08 & {\fontsize{9.5pt}{12.5pt}\selectfont \textbf{181.29}}  & 0.199 & 3.96 & 0.025 & 0.870 \\
        & PID & 157.80 & 254.05 & 0.177 & 3.40 & 0.019 & 0.770 \\
        & Invariant-set & 151.23 & 186.43 & {\fontsize{9.5pt}{12.5pt}\selectfont    \textbf{0.001}} & {\fontsize{9.5pt}{12.5pt}\selectfont   \textbf{0.892}}  & {\fontsize{9.5pt}{12.5pt}\selectfont  \textbf{0.011}}  & {\fontsize{9.5pt}{12.5pt}\selectfont  \textbf{0.542}}  \\
        \hline
    \end{tabular}
    }
\end{table*}

\begin{table*}[t]
    \centering
    \caption{Performance Metrics of controllers on Figure-Of-Eight shape curve}
    \label{tab:fig_8_metrics}
    \renewcommand{\arraystretch}{1.5}
    \footnotesize

    \resizebox{\textwidth}{!}{%
    \begin{tabular}{p{2.0cm} | r p{2.5cm} r r r r}
        \textbf{Controller} 
        & \multicolumn{6}{c}{\textbf{Controller Metrics}} \\
        \cline{2-7}
        & \textbf{Time Taken(s)} & \textbf{Obstacle Proximity at Crossing (m)} 
        & \textbf{Avg Accel.} & \textbf{Max Accel.} & \textbf{Avg Curv.} & \textbf{Max Curv.} \\
        \hline
         MPC  & - & - & - & - & - & - \\
         PID & 724.6 & 25.75 & 0.01 & 2.81 & 0.049 & 0.19 \\
         Invariant-set & 1054 & 3.81 & 0.025  & 0.43  & 0.015 & 0.57  \\
        \hline
    \end{tabular}
    }
\end{table*}

\begin{figure}[ht]
            \centering
            \includegraphics[width=0.45\textwidth]{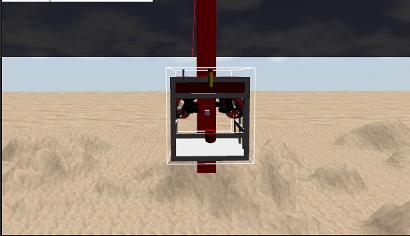}
            \caption{Scenario 4 : Underwater Environment with a infinitely long static column ahead of the vehicle.}
            \label{fig:long}
\end{figure}

\textbf{Scenario 4} : This scenario shown in Figure~\ref{fig:long}, presents an underwater environment where the vehicle encounters a vertically extended column-like obstacle situated directly along its intended path. Unlike a wide or expansive barrier, this obstruction is narrow in width but extends vertically across the depth range, requiring the vehicle to find a lateral route around it. This setup reflects practical situations such as underwater pillars, mooring structures or subsea installations that are tall but confined in horizontal spread. The challenge lies in detecting the obstruction early and executing precise lateral maneuvering while maintaining depth and orientation. The navigation goal is positioned 110 meters straight ahead at the same depth, making this scenario an important benchmark for evaluating a UUV’s ability to perform accurate lateral avoidance while preserving trajectory stability in the presence of persistent vertical impediments.

\begin{figure}[ht]
            \centering
            \includegraphics[width=0.45\textwidth]{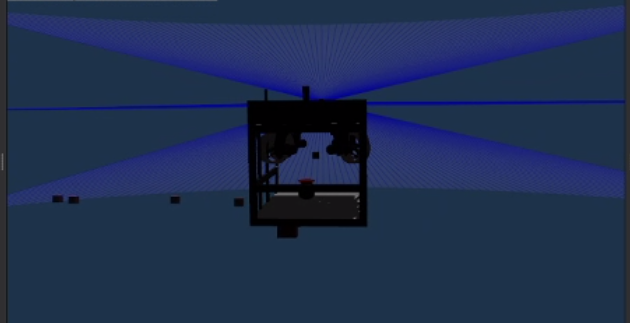}
            \caption{Scenario 5 : Underwater Environment with Temporally Varying Dynamic Obstacles in the Vehicle’s Field of View .}
            \label{fig:dyn}
\end{figure}

\textbf{Scenario 5} : In this scenario illustrated in Figure~\ref{fig:dyn}, the underwater vehicle navigates an environment populated with dynamic obstacles that intermittently enter and exit its field of view. These moving objects, which may simulate marine life, drifting debris, or other transient underwater elements, introduce unpredictability and demand real-time situational awareness. The objective of this scenario is to rigorously evaluate the vehicle’s ability to make optimal and robust decisions under uncertainty—adjusting its trajectory on the fly to avoid collisions while maintaining progress toward a predefined goal located ahead. It serves as a critical test of the system's reactive planning and control capabilities, particularly in scenarios where static path assumptions fail and continuous adaptation is required for safe and effective navigation.

\begin{figure}[ht]
            \centering
            \includegraphics[width=0.50\textwidth]{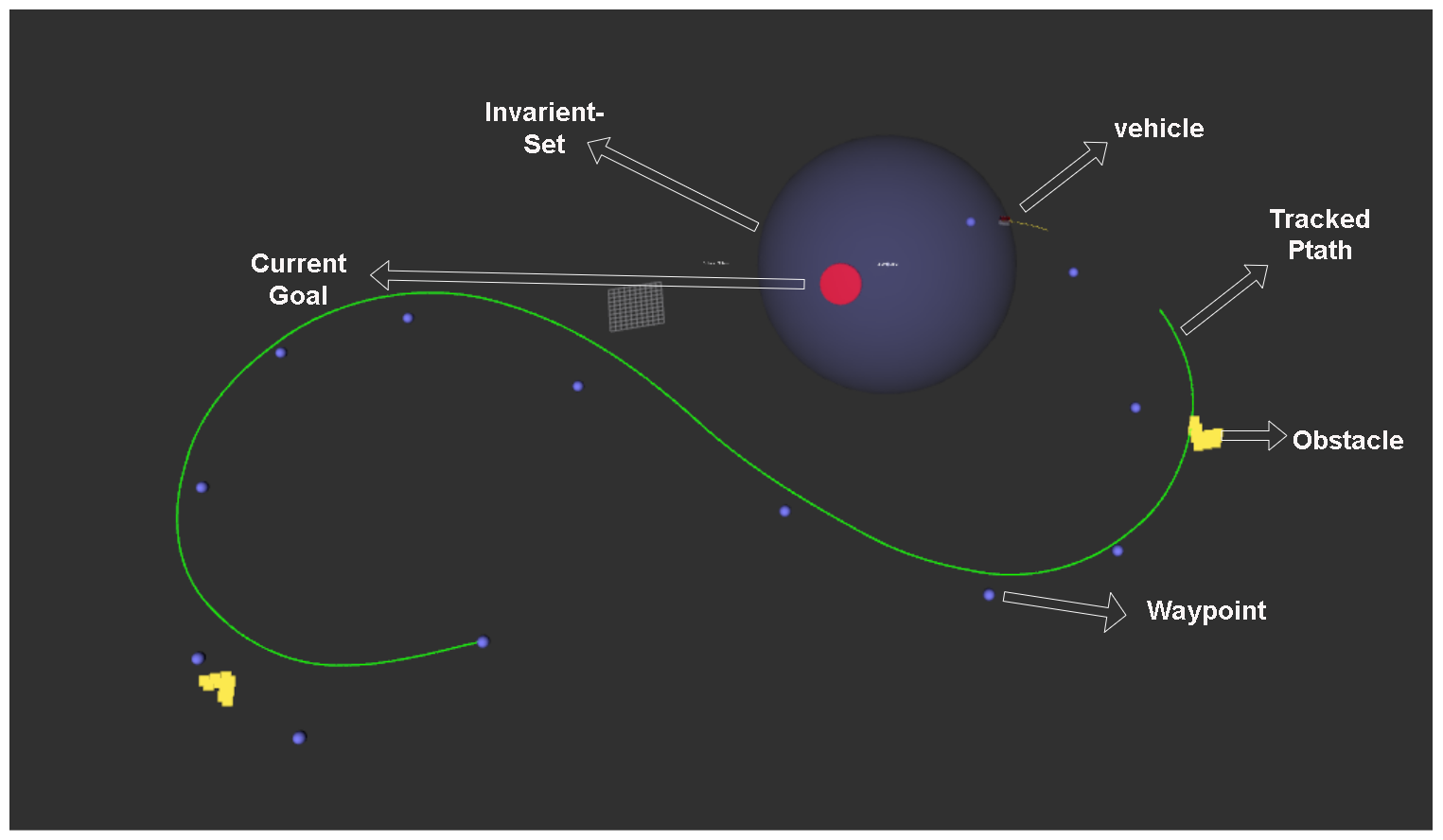}
            \caption{Tracked Path visualization of Figure-Of-Eight curve using Invariant-set controller}
            \label{fig:8_ipc}
\end{figure}

\begin{figure}[ht]
            \centering
            \includegraphics[width=0.50\textwidth]{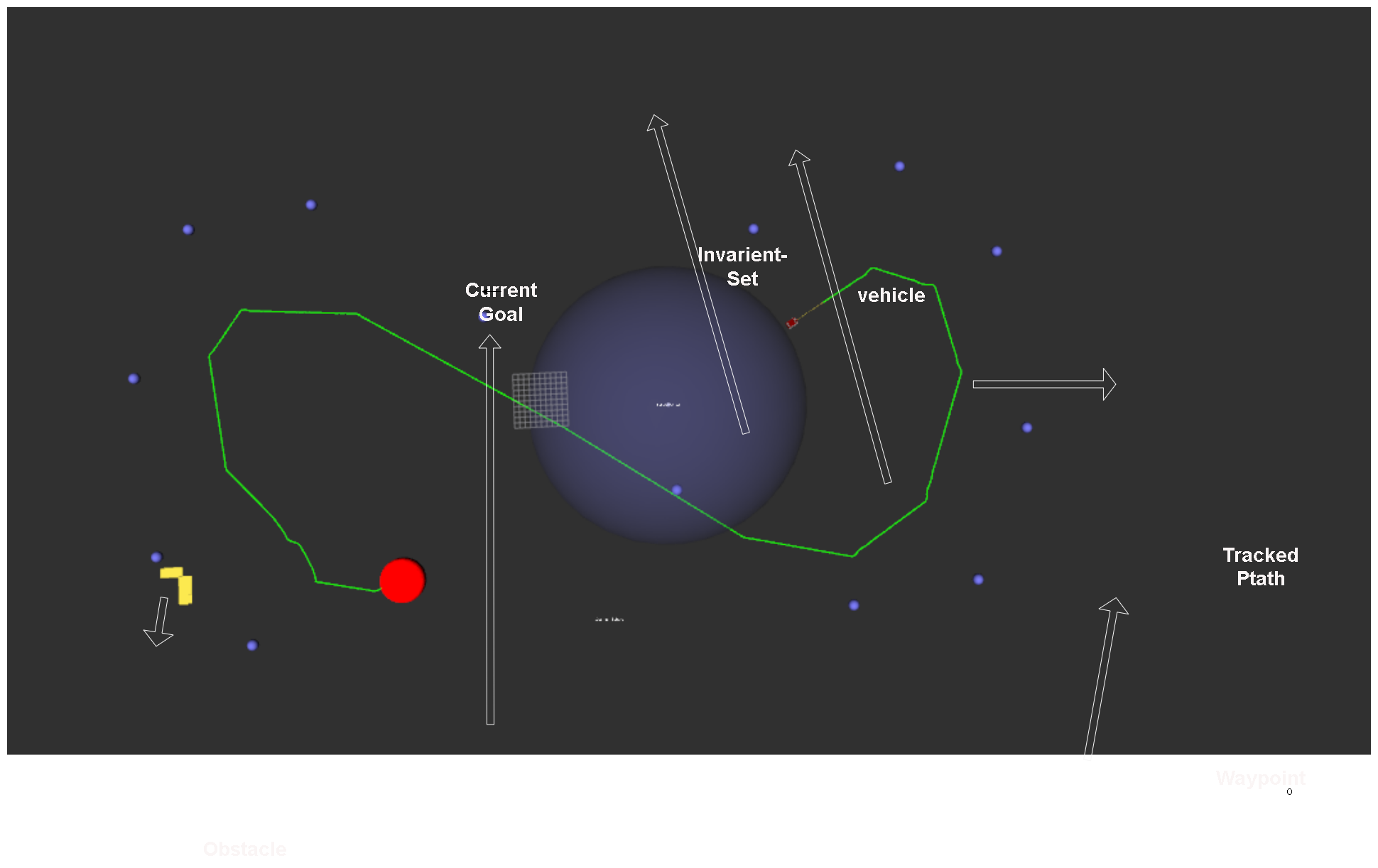}
            \caption{Tracked Path visualization of Figure-Of-Eight curve using PID controller}
            \label{fig:8_pid}
\end{figure}

Next, we evaluate the performance of coupled methods for sensor based planning and control under the listed scenarios. 
In Table ~\ref{tab:performance_metrics}, we show various controllers used in conjunction with the integrated planner across different scenarios to highlight their performance and behavior. Apart from tracking accuracy, the time taken to reach the goal is an important metric. Controllers like MPC consistently yield lower times, reflecting better responsiveness and constraint handling. In contrast, PID tends to lag, and invariant-set based on (used in IPC) shows scenario-dependent timing influenced by curvature complexity. Meanwhile, acceleration further reflects control effort and efficiency. The MPC reflects balanced max and average profile, PID results into higher peaks, suggesting more aggressive and potentially unstable responses. Eventually, invariant-set based controller tends to minimize average acceleration profile, indicate smooth and less energy consuming behavior. Lastly, path curvature provides insights into maneuverability demands. MPC and Invariant-set based controller maintain lower maximum and average curvature, which is beneficial for vehicles with limited turning capabilities. In contrast, PID sometimes induces sharp turns, increasing the control burden and potentially compromising trajectory smoothness.

The IPC planner is selected as the baseline because it functions as a sensor-based planner, offering a reliable and reactive approach for safe and efficient navigation \cite{Karthik2021, Karthik2024}. 

To assess the system’s robustness more thoroughly, a complex figure-eight trajectory was designed with obstacles strategically placed along the path. A threshold radius of 30 meters was defined around each goal point—once the vehicle enters this region, the goal automatically shifts to the next point in the sequence.
This test was designed to assess the system’s ability to maintain stability, handle tight turns, respect the inter-dependencies of a coupled control system, avoid obstacles, and accurately follow the intended path. In this setting, the invariant set-based controller performed well, demonstrating smooth tracking and consistent stability. MPC, however, struggled with the complexity of the path—it was unable to handle the sharp turns effectively, leading to instability and eventual failure to complete the trajectory.


\subsubsection{Invariant-set based Controller }
\begin{table*}[!t] 
\centering 
\caption{Comparison of Controllers for UUV Navigation}
\label{tab:controller_comparison}

\newlist{compactitem}{itemize}{1}
\setlist[compactitem]{label=\textbullet, nosep, leftmargin=*, topsep=0pt, partopsep=0pt}

\begin{tabular*}{\textwidth}{@{\extracolsep{\fill}} p{0.07\textwidth} p{0.22\textwidth} p{0.3\textwidth} p{0.16\textwidth} p{0.18\textwidth} @{}} 
\toprule
\textbf{Controller} & \textbf{Key Advantages} & \textbf{Key Limitations} & \textbf{Suitable Scenarios} & \textbf{Less Suitable Scenarios} \\
\midrule

\textbf{PID} &
\begin{compactitem}
    \item Simplicity and ease of implementation.
    \item Effective in linear environments with minimal oscillations.
\end{compactitem} &
\begin{compactitem}
    \item Lacks predictive capabilities; struggles with sudden changes.
    \item Higher acceleration peaks and sharp turns.
    \item No inherent constraint handling or obstacle avoidance.
\end{compactitem} &
\begin{compactitem}
    \item Linear path following.
    \item Simple waypoint tracking.
\end{compactitem} &
\begin{compactitem}
    \item Dynamic obstacle avoidance.
    \item Complex maneuvering.
    \item Highly constrained environments.
\end{compactitem} \\
\midrule

\textbf{MPC} &
\begin{compactitem}
    \item Time-effective performance and path optimization.
    \item Reduces deviations and maintains goal bias.
    \item Strong performance in narrow passages by balancing optimality and local constraints.
\end{compactitem} &
\begin{compactitem}
    \item Less responsive to rapidly changing environments if prediction horizon is mismatched.
    \item Highly sensitive to operating conditions and velocity range.
    \item Challenges with coupled control dynamics; poor performance in tightly coupled maneuvers (e.g., Figure-Of-Eight), experiencing significant difficulties in completing the Figure-Of-Eight trajectory and exhibiting instability.
\end{compactitem} &
\begin{compactitem}
    \item General obstacle avoidance.
    \item Path optimization in somewhat predictable dynamic environments.
\end{compactitem} &
\begin{compactitem}
    \item Highly dynamic environments with rapid changes.
    \item Complex multi-axis maneuvers, scenarios where tuning is difficult.
\end{compactitem} \\
\midrule

\textbf{Invariant-set} &
\begin{compactitem}
    \item Provides strong safety guarantees by design and ensures adherence to operational constraints.
    \item Superior stability and performance in complex, tightly coupled maneuvers (e.g., Figure-Of-Eight).
    \item Smooth control profile with low average acceleration and curvature.
\end{compactitem} &
\begin{compactitem}
    \item Intrinsically imposes a minimum turning radius, leading to fixed minimum invariant set size.
    \item Can be inefficient or unsuitable in extremely constrained spaces (e.g., avoids very narrow passages).
    \item May lead to non-optimal behavior from mission objectives in specific constrained tasks.
\end{compactitem} &
\begin{compactitem}
    \item Missions prioritizing safety and guaranteed constraint adherence.
    \item Complex and continuous maneuvers.
    \item Highly dynamic and cluttered environments.
\end{compactitem} &
\begin{compactitem}
    \item Tasks requiring navigation through extremely narrow passages.
    \item Scenarios where strict path optimality (shortest path) is the sole objective.
\end{compactitem} \\
\bottomrule
\end{tabular*}

\end{table*}
\textbf{Advantages :} Table ~\ref{tab:performance_metrics} demonstrates that the invariant-set framework based on invariant set performs conservatively yet effectively, producing the most efficient trajectory within a reasonable time when compared to the other approaches. The Invariant set \cite{Karthik2024} based controller closely follows the desired trajectory, resulting in minimal path length. Moreover, it offers the flexibility to incorporate vehicle-specific operational constraints such as turning radius, yaw, surge and pitch rates, ensuring that the vehicle remains within a defined threshold region around the goal. This controller exhibits superior performance in both dynamic and cluttered environment, where responsiveness and adaptability are critical. This robustness represents lower tracking error and smooth control profile under obstacle rich environment. Additionally, it's performance across the figure-of-eight figure ~\ref{tab:fig_8_metrics} shows its capabilities as an overall integrated system well suited to work under challenging environments. It successfully tracks down all the given points, maintaining a smooth trajectory ~\ref{fig:8_ipc} with lower angular curvature.  \\
\textbf{Disadvantages :} The Invariant-set framework intrinsically imposes a minimum turning radius, leading to the continual generation of spherical invariant sets with a fixed minimum size. While this is effective in general scenarios, it becomes inefficient or unsuitable in environments where turning is unnecessary or the space is too constrained. For instance, in the narrow passage scenario (Figure ~\ref{fig:narrow}) discussed above, the invariant set based controller avoid entering the passage entirely, as its width is smaller than the required minimum radius of the invariant set. Instead, the vehicle pitches upward, likely interpreting the passage entrance as an obstacle, bypasses the narrow region, and then pitches downward to re-align with and reach the goal. This behavior can lead to unintended consequences, such as non-optimal behavior or deviation from mission objectives, particularly in tasks like pipeline inspection where navigating through narrow spaces and staying close to structures is essential. \\

\subsubsection {PID Controller}

\textbf{Advantages :} Although PID control is the most fundamental approach to path tracking, it remains reliable by adhering strictly to its core principle of minimizing error relative to the target. In the context of intermediate waypoint navigation, the PID controller consistently attempts to converge toward the assigned goal points without over- complicating the control structure, a fact clearly supported by the results presented in Tables~\ref{tab:performance_metrics} and~\ref{tab:fig_8_metrics}. Across various test scenarios, it has demonstrated stable and predictable behavior. It particularly performed better in environment with long linear barrier, where its steady state characteristics helps to maintain directionality with minimal oscillations.  Even in the more complex figure-eight trajectory, while it may not match the smoothness of the invariant set controller, the PID controller successfully tracked all waypoints and maintained system stability (Figure~\ref{fig:8_pid}). Due to its inherent simplicity and robustness, it is an excellent candidate for serving as a secondary error correction mechanism—complementing more complex controllers, to minimize deviation and enhance overall tracking precision.\\
\textbf{Disadvantages :}Since PID control does not optimize trajectories, it often results in suboptimal performance. This manifests as longer path lengths and increased time to reach the target compared to more advanced control strategies. Furthermore, relying solely on PID can be limiting in complex or dynamic environments, where reactive obstacle avoidance and adherence to vehicle constraints are critical. Without predictive capabilities, PID may struggle with sudden environmental changes, leading to inefficient maneuvers or failure to reach the goal safely. Therefore, while useful for error correction and low-level control, the PID is best employed in conjunction with higher-level planners or model-based controllers.\\

\subsubsection { MPC Controller}

\textbf{Advantages :} This combination has yielded the most time-effective performance across the test cases discussed above. The Model Predictive Control  (MPC) (\cite{gomes2018}) framework plays a critical role in this outcome by continuously optimizing the robot’s trajectory over a prediction horizon. It not only reduces unnecessary deviations or drifts from the intended path but also maintains a consistent bias toward reaching the global goal. By factoring in both the immediate environment and the long-term objective, MPC ensures that the system progresses safely and efficiently. As a result,

The goal is achieved in the most optimal manner possible, balancing short-term responsiveness with long-term planning for both performance and safety. This combination has shown strong performance in narrow passage scenarios by closely following the channel, even though the integrated planner imposes a constraint on the minimum radius of the invariant sphere due to its turning radius limitations. It also demonstrate robust performance in cluttered and wide obstacle environment, where MPC adjusts trajectories to maintain stability and minimize deviations.\\
\textbf{Disadvantages :} Several limitations were identified during the implementation of the Model Predictive Control (MPC) strategy. Compared to the invariant set-based control approach, MPC was notably less responsive to rapidly changing environments. This is largely due to the fact that MPC generates a control trajectory over a fixed prediction horizon using the current goal. When the frequency of local goal updates is higher than the prediction horizon's responsiveness, the controller may fail to adapt quickly enough. As a result, it can exhibit delayed or entirely missed control responses, leading to significant deviations from the intended path—and in some cases, preventing the robot from reaching the goal.

The performance metrics cited were recorded under the constraint of a maximum linear velocity of 2 m/s and an angular velocity of 0.025 rad/s. While other control algorithms demonstrated consistent and reliable behavior across a range of both low and high speeds, the MPC controller proved to be far more sensitive to operating conditions. It functioned reliably only within a narrow band of velocity values, making it unsuitable for more dynamic or unpredictable scenarios. These limitations make the MPC approach highly restrictive.

Another critical drawback lies in the controller's heavy dependence on configuration parameters, particularly the prediction horizon length. The horizon significantly influences the controller’s adaptability, responsiveness, and accuracy. If not appropriately tuned, the internal optimiser can fail to converge on a viable solution, effectively causing the system to get stuck in suboptimal or non-functional control loops. In such cases, rather than progressing toward the goal, the optimiser may spiral into an internal loop without generating effective outputs, resulting in inefficient or failed navigation. 

Additionally, MPC faces challenges when dealing with coupled control dynamics, for example, in our system, the control of pitch, yaw, and roll is interdependent—changes in one axis influence the others. Specifically, a change in pitch affects both roll and yaw, requiring coordinated compensation to maintain stability. However, MPC, in its basic form, does not inherently manage such coupling unless the dynamic relationships are explicitly modeled and well understood. This makes the tuning and implementation process both complex and labor-intensive.

Due to these limitations, MPC performed poorly in scenarios with tightly coupled maneuvers—such as the figure-eight path. It failed to consistently track the sequence of goal points, and after two or three transitions, the system would become unstable and eventually crash.

That said, despite these limitations, MPC—much like PID control—still holds potential as a low-level controller when paired with a high-level planning or supervisory system. In such an architecture, the high-level controller could oversee global trajectory planning and decision-making, while MPC manages smooth, localized execution of control actions. This hierarchical approach can improve both the responsiveness and robustness of autonomous systems operating in complex environments.

\subsubsection {Summary}

The comparative analysis in Table ~\ref{tab:controller_comparison} highlights distinct characteristics among UUV navigation controllers. The PID controller offers straightforward implementation and effectiveness in linear environments, proving suitable for simple waypoint tracking. However, its lack of predictive capacity renders it ill-suited for dynamic obstacle avoidance or complex, sudden maneuvers, often resulting in sharp turns and high acceleration peaks.

Conversely, the Model Predictive Control (MPC) demonstrates time-effective path optimization and deviation reduction, making it generally effective for obstacle avoidance in somewhat predictable dynamic settings. Its limitations surface in rapidly changing or tightly coupled multi-axis maneuvers, like a Figure-Of-Eight trajectory, where sensitivity to operating conditions can lead to instability.

The Invariant-set controller distinguishes itself with inherent safety guarantees and superior stability, particularly excelling in complex, continuous maneuvers and cluttered dynamic environments. It provides a smooth control profile with low acceleration. A key limitation, however, is its imposed minimum turning radius, which can make it inefficient or unsuitable for extremely narrow passages where strict path optimality is the sole objective. Ultimately, the optimal controller choice depends on the specific mission's priorities, balancing safety, maneuverability, and environmental dynamics.

\section{Conclusion and Future Scope}\label{sec6}

This survey presents a comprehensive analysis of sensor-based planning and control strategies employed for Unmanned Underwater Vehicles (UUVs) operating in complex and uncertain underwater environments. The study introduces a taxonomy of existing methodologies, broadly categorizing them into decoupled and coupled architectures, based on the degree of integration between the planning and control layers.

In contrast, coupled architectures implement a tighter integration of planning and control, where sensor feedback directly influences both the trajectory generation and the associated control actions in a closed-loop fashion. Such approaches are particularly well-suited for real-time adaptation in environments characterized by poor visibility, variable currents, moving obstacles, and sensor noise. The survey identifies that coupled strategies inherently offer better safety and responsiveness, making them advantageous for navigation in highly dynamic or uncertain operational scenarios. However, the choice of controller within these systems presents critical trade-offs: for instance, while Model Predictive Control (MPC) excels at path optimization, it can be computationally demanding, whereas Invariant-set controllers provide robust safety guarantees at the cost of maneuverability, and simpler PID controllers often lack the predictive power for complex obstacle avoidance. Coupled architectures show strong potential as unified solutions for diverse underwater navigation challenges, warranting further in-depth investigation.

While decoupled techniques have advanced significantly and provide a strong theoretical foundation, the full potential of coupled sensor-based approaches remains underexplored—especially in terms of achieving simultaneous optimality, safety, and robustness guarantees. The survey advocates for scenario-driven benchmarking to evaluate these methods under realistic mission constraints and calls for further research into formalizing stability and performance bounds for coupled frameworks. This direction is critical to enabling next-generation UUVs to achieve autonomous, resilient, and mission-aware behavior in the absence of global positioning and with limited communication capabilities.

\section*{Acknowledgments}

 We acknowledge the Centre for Artificial Intelligence and Robotics (CAIR) at Bengaluru, DRDO, India for their research grants to carry out this research (Ref. no:  RD/0122-DRDO010-003).  

\section*{Financial disclosure}

None reported.

\section*{Conflict of interest}

The authors declare no potential conflict of interests.


\bibliographystyle{IEEEtran}

\bibliography{ZwileyNJD-AMA}

@inproceedings{cohen2023,
  title={Set-transformer BeamsNet for {AUV} velocity forecasting in complete {DVL} outage scenarios},
  author={Cohen, Nadav and Yampolsky, Zeev and Klein, Itzik},
  booktitle={2023 IEEE Underwater Technology (UT)},
  pages={1-6},
  year={2023},
  organization={IEEE}
}

@inproceedings{osse2007,
  title={The Deepglider: A full ocean depth glider for oceanographic research},
  author={Osse, T James and Eriksen, Charles C},
  booktitle={OCEANS 2007},
  pages={1-12},
  year={2007},
  organization={IEEE}
}

@inproceedings{vestgard2000,
  title={{HUGIN} 3000 {AUV} for deepwater surveying},
  author={Vestgard, Karstein and Klepaker, Rolf Arne and Storkersen, Nils},
  booktitle={Offshore Technology Conference},
  pages={OTC--12005},
  year={2000},
  organization={OTC}
}

@incollection{gary2008,
  title={{3D} mapping and characterization of Sistema Zacat{\'o}n from {DEPTHX (Deep  Hreatic Thermal Explorer)}},
  author={Gary, Marcus and Fairfield, Nathaniel and Stone, William C and Wettergreen, David and Kantor, George and Sharp, Jr, John M},
  booktitle={Sinkholes and the Engineering and Environmental Impacts of Karst},
  pages={202--212},
  year={2008}
}

@inproceedings{stokey2005,
  title={Development of the {REMUS 600} autonomous underwater vehicle},
  author={Stokey, Roger P and Roup, Alexander and von Alt, Chris and Allen, Ben and Forrester, Ned and Austin, Tom and Goldsborough, Rob and Purcell, Mike and Jaffre, Fred and Packard, Greg and others},
  booktitle={Proceedings of OCEANS 2005 {MTS/IEEE}},
  pages={1301--1304},
  year={2005},
  organization={IEEE}
}

@inproceedings{williams2000,
  title={Autonomous underwater simultaneous localisation and map building},
  author={Williams, Stefan B and Newman, Paul and Dissanayake, Gamini and Durrant-Whyte, Hugh},
  booktitle={Proceedings 2000 {ICRA}.},
  volume={2},
  pages={1793--1798},
  year={2000},
  organization={IEEE}
}

@article{mallios2014, 
  title={Scan matching {SLAM} in underwater environments},
  author={Mallios, Angelos and Ridao, Pere and Ribas, David and Hern{\'a}ndez, Emili},
  journal={Autonomous Robots},
  volume={36},
  pages={181--198},
  year={2014},
  publisher={Springer}
}

@inproceedings{wang2015,
  title={Underwater electric current communication of robotic fish: Design and experimental results},
  author={Wang, Wei and Zhao, Jianwei and Xiong, Wei and Cao, Fayang and Xie, Guangming},
  booktitle={2015 IEEE International Conference on Robotics and Automation ({ICRA})},
  pages={1166--1171},
  year={2015},
  organization={IEEE}
}

@article{keerthi2020,
  title={Into the world of underwater swarm robotics: Architecture, communication, applications and challenges},
  author={Keerthi, Koyippilly S and Mahapatra, Bandana and Menon, Varun Girijan},
  journal={Recent Advances in Computer Science and Communications (Formerly: Recent Patents on Computer Science)},
  volume={13},
  number={2},
  pages={110--119},
  year={2020},
  publisher={Bentham Science Publishers}
}

@article{chin2018,
  title={Robust genetic algorithm and fuzzy inference mechanism embedded in a sliding-mode controller for an uncertain underwater robot},
  author={Chin, Cheng Siong and Lin, Wei Peng},
  journal={{IEEE/ASME} transactions on mechatronics},
  volume={23},
  number={2},
  pages={655--666},
  year={2018},
  publisher={IEEE}
}

@inproceedings{ishibashi2006,
  title={The Improvement of the Prescision of an Inertial Navigation System for {AUV} based on the Neural Network},
  author={Ishibashi, Shojiro},
  booktitle={OCEANS 2006-Asia Pacific},
  pages={1--6},
  year={2006},
  organization={IEEE}
}

@article{zhang2020,
  title={{NavNet}: {AUV} navigation through deep sequential learning},
  author={Zhang, Xin and He, Bo and Li, Guangliang and Mu, Xiaokai and Zhou, Ying and Mang, Tanji},
  journal={{IEEE Access}},
  volume={8},
  pages={59845--59861},
  year={2020},
  publisher={IEEE}
}

@article{li2018,
  title={Path planning technologies for autonomous underwater vehicles-a review},
  author={Li, Daoliang and Wang, Peng and Du, Ling},
  journal={{IEEE Access}},
  volume={7},
  pages={9745--9768},
  year={2018},
  publisher={IEEE}
}

@conference{Stone2006,
author = {William Stone and Nathaniel Fairfield and George A. Kantor},
title = {Autonomous Underwater Vehicle Navigation and Proximity Operations for Deep Phreatic Thermal Explorer ({DEPTHX})},
booktitle = {Proceedings of Masterclass in {AUV} Technology for Polar Science},
year = {2006},
month = {March},
editor = {Griffiths, G. and Collins, K.},
publisher = {Society for Underwater Technology},
address = {London},
}

@ARTICLE{Wynn2014,
  author={Russell B. Wynn and Veerle A.I. Huvenne and Timothy P. Le Bas and Bramley J. Murton and Douglas P. Connelly and Brian J. Bett and Henry A. Ruhl and Kirsty J. Morris and Jeffrey Peakall and Daniel R. Parsons and Esther J. Sumner and Stephen E. Darby and Robert M. Dorrell and James E. Hunt},
  journal={Marine Geology},
  title={Autonomous Underwater Vehicle ({AUV}): Their past, present and future contributions to the advancement of marine geoscience},
  journal={Marine geology},
  year={2014},
  volume={352},
  pages={451-468}
}

@article{Sahl2010,
author = {Sahl, Jason W. and Gary, Marcus O. and Harris, J. Kirk and Spear, John R.},
title = {A comparative molecular analysis of water-filled limestone sinkholes in north-eastern Mexico},
journal = {Environmental Microbiology},
volume = {13},
number = {1},
pages = {226-240},
eprint = {https://enviromicro-journals.onlinelibrary.wiley.com/doi/pdf/10.1111/j.1462-2920.2010.02324.x},
year = {2010}
}

@article{Gafurov2015,
title = {Autonomous Unmanned Underwater Vehicles Development Tendencies},
journal = {Procedia Engineering},
volume = {106},
pages = {141-148},
year = {2015},
note = {Proceedings of the 2nd International Conference on Dynamics and Vibroacoustics of Machines (DVM2014) September 15 –17, 2014 Samara, Russia},
issn = {1877-7058},
author = {Salimzhan A. Gafurov and Evgeniy V. Klochkov},
}

@INPROCEEDINGS{Nodland1981,
  author={Nodland, W. and Ewart, T. and Bendiner, W. and Miller, J. and Aagaard, E.},
  booktitle={OCEANS 81}, 
  title={{SPURV II} An Unmanned, Free-Swimming Submersible Developed for Oceanographic Research}, 
  year={1981},
  volume={},
  number={},
  pages={92-98},
  doi={10.1109/OCEANS.1981.1151607}}

@article{Paull2012 ,
  title={Sensor-driven online coverage planning for autonomous underwater vehicles},
  author={Paull, Liam and Saeedi, Sajad and Seto, Mae and Li, Howard},
  journal={IEEE/ASME Transactions on Mechatronics},
  volume={18},
  number={6},
  pages={1827--1838},
  year={2012},
  publisher={IEEE}
}

@article{Yang2021 ,
  title={A survey of autonomous underwater vehicle formation: Performance, formation control, and communication capability},
  author={Yang, Yue and Xiao, Yang and Li, Tieshan},
  journal={IEEE Communications Surveys \& Tutorials},
  volume={23},
  number={2},
  pages={815--841},
  year={2021},
  publisher={IEEE}
}

@article{Budiyono2009,
  title={Advances in unmanned underwater vehicles technologies: Modeling, control and guidance perspectives},
  author={Budiyono, Agus},
  year={2009},
  journal = {IEEE Transactions on Robotics},
  publisher={CSIR}
}

@article{Balestrieri2021,
  title={Sensors and measurements for unmanned systems: An overview},
  author={Balestrieri, Eulalia and Daponte, Pasquale and De Vito, Luca and Lamonaca, Francesco},
  journal={Sensors},
  volume={21},
  number={4},
  pages={1518},
  year={2021},
  publisher={MDPI}
}

@article{Crasta2018,
  title={Multiple autonomous surface vehicle motion planning for cooperative range-based underwater target localization},
  author={Crasta, Naveena and Moreno-Salinas, David and Pascoal, Ant{\'o}nio M and Aranda, Joaqu{\'\i}n},
  journal={Annual Reviews in Control},
  volume={46},
  pages={326--342},
  year={2018},
  publisher={Elsevier}
}

@article{Zacchini2022,
  title={Sensor-driven autonomous underwater inspections: A receding-horizon {RRT-based} view planning solution for {AUV}},
  author={Zacchini, Leonardo and Franchi, Matteo and Ridolfi, Alessandro},
  journal={Journal of Field Robotics},
  volume={39},
  number={5},
  pages={499--527},
  year={2022},
  publisher={Wiley Online Library}
}

@article{Valavanis1997,
  title={Control architectures for autonomous underwater vehicles},
  author={Valavanis, Kimon P and Gracanin, Denis and Matijasevic, Maja and Kolluru, Ramesh and Demetriou, Georgios A},
  journal={{IEEE} Control Systems Magazine},
  volume={17},
  number={6},
  pages={48--64},
  year={1997},
  publisher={IEEE}
}

@article{Neira2021,
  title={Review on unmanned underwater robotics, structure designs, materials, sensors, actuators, and navigation control},
  author={Neira, Javier and Sequeiros, Cristhel and Huamani, Richard and Machaca, Elfer and Fonseca, Paola and Nina, Wilder},
  journal={Journal of Robotics},
  volume={2021},
  number={1},
  pages={5542920},
  year={2021},
  publisher={Wiley Online Library}
}

@article{Melo2017,
  title={Survey on advances on terrain based navigation for autonomous underwater vehicles},
  author={Melo, Jos{\'e} and Matos, An{\'\i}bal},
  journal={Ocean Engineering},
  volume={139},
  pages={250--264},
  year={2017},
  publisher={Elsevier}
}

@article{Felemban2015,
  title={Underwater sensor network applications: A comprehensive survey},
  author={Felemban, Emad and Shaikh, Faisal Karim and Qureshi, Umair Mujtaba and Sheikh, Adil A and Qaisar, Saad Bin},
  journal={International Journal of Distributed Sensor Networks},
  volume={11},
  number={11},
  pages={896832},
  year={2015},
  publisher={SAGE Publications Sage UK: London, England}
}

@article{Antonelli2001,
  title={Real-time path planning and obstacle avoidance for {RAIS}: an autonomous underwater vehicle},
  author={Antonelli, Gianluca and Chiaverini, Stefano and Finotello, Roberto and Schiavon, Riccardo},
  journal={IEEE Journal of Oceanic Engineering},
  volume={26},
  number={2},
  pages={216--227},
  year={2001},
  publisher={IEEE}
}

@article{Alamdari2020,
  title={Robust trajectory tracking control for underactuated autonomous underwater vehicles in uncertain environments},
  author={Heshmati-Alamdari, Shahab and Nikou, Alexandros and Dimarogonas, Dimos V},
  journal={IEEE Transactions on Automation Science and Engineering},
  volume={18},
  number={3},
  pages={1288--1301},
  year={2020},
  publisher={IEEE}
}

@article{Bovio2006,
  title={Autonomous underwater vehicles for scientific and naval operations},
  author={Bovio, Edoardo and Cecchi, D and Baralli, Francesco},
  journal={Annual Reviews in Control},
  volume={30},
  number={2},
  pages={117--130},
  year={2006},
  publisher={Elsevier}
}

@article{Bennett2000,
  title={A behavior-based approach to adaptive feature detection and following with autonomous underwater vehicles},
  author={Bennett, Andrew A and Leonard, John J},
  journal={IEEE Journal of Oceanic Engineering},
  volume={25},
  number={2},
  pages={213--226},
  year={2000},
  publisher={IEEE}
}

@article{Petillot2019,
  title={Underwater robots: From remotely operated vehicles to intervention-autonomous underwater vehicles},
  author={Petillot, Yvan R and Antonelli, Gianluca and Casalino, Giuseppe and Ferreira, Fausto},
  journal={IEEE Robotics \& Automation Magazine},
  volume={26},
  number={2},
  pages={94--101},
  year={2019},
  publisher={IEEE}
}

@incollection{Leonard2013,
  author = {John J. Leonard and Alexander Bahr},
  title = {Autonomous Underwater Vehicle Navigation},
  booktitle = {Marine Robot Autonomy},
  publisher = {Springer, New York},
  year = {2013},
  pages = {341-355}
}

@article{Williams2001,
  author = {Stefan B. Williams and Paul Newman and Julio Rosenblatt and Gamini Dissanayake and Hugh Durrant-Whyte},
  title = {Autonomous underwater navigation and control},
  journal = {Robotica},
  volume = {19},
  number = {5},
  pages = {481-496},
  year = {2001},
  publisher = {Cambridge University Press}
}

@article{Miller2010,
  author = {Paul A. Miller and Jay A. Farrell and Yuanyuan Zhao and Vladimir Djapic},
  title = {Autonomous Underwater Vehicle Navigation},
  journal = {IEEE Journal of Oceanic Engineering},
  volume = {35},
  number = {3},
  pages = {663-675},
  year = {2010},
  publisher = {IEEE}
}

@article{Bobkov2018,
  author = {V. A. Bobkov and A. P. Kudryashov and S. V. Mel'man and A. F. Shcherbatyuk},
  title = {Autonomous Underwater Navigation with {3D} Environment Modeling Using Stereo Images},
  journal = {Gyroscopy and Navigation},
  volume = {9},
  number = {1},
  pages = {67-75},
  year = {2018},
  publisher = {Pleiades Publishing Ltd.}
}

@inproceedings{Jalal2021,
  author = {Fahad Jalal and Faizan Nasir},
  title = {Underwater Navigation, Localization and Path Planning for Autonomous Vehicles: A Review},
  booktitle = {2021 International Bhurban Conference on Applied Sciences and Technologies ({IBCAST})},
  pages = {1-8},
  year = {2021},
  organization = {IEEE}
}

@inproceedings{Panish2011,
  title={Achieving high navigation accuracy using inertial navigation systems in autonomous underwater vehicles},
  author={Panish, Robert and Taylor, Mikell},
  booktitle={OCEANS 2011 IEEE-Spain},
  pages={1--7},
  year={2011},
  organization={IEEE}
}

@article{Yu2001,
  title={Navigation of autonomous underwater vehicles based on artificial underwater landmarks},
  author={Yu, Son-Cheol and Ura, Tamaki and Fujii, Teruo and Kondo, Hayato},
  booktitle={MTS/IEEE Oceans 2001. An Ocean Odyssey. Conference Proceedings (IEEE Cat. No. 01CH37295)},
  volume={1},
  pages={409--416},
  year={2001},
  organization={IEEE}
}

@article{Yao2019,
  title={Review of path planning for autonomous underwater vehicles},
  author={Yao, Tingting and He, Tao and Zhao, WenLong and M. Sani, Abdou Yahouza},

  jornal = {Proceedings of the 2019 International Conference on Robotics, Intelligent Control and Artificial Intelligence},

  pages={482--487},
  year={2019}
}

@article{Paull2014,
  author = {Liam Paull and Sajad Saeedi and Mae Seto and Howard Li},
  title = {{AUV} Navigation and Localization: A Review},
  journal = {IEEE Journal of Oceanic Engineering},
  volume = {39},
  number = {1},
  pages = {131--145},
  year = {2014},
  publisher = {IEEE}
}

@inproceedings{Zhe2022,
  author = {Zhe Zhang and Shusi Chen and Yanghui Li and Lifeng Wang and Rui Ren and Lei Xu and Jun Wang and Xianchun Zhang},
  title = {Local Path Planning of Unmanned Underwater Vehicle Based on Improved {APF} and Rolling Window Method},
  booktitle = {2022 International Conference on Cyber-Physical Social Intelligence ({ICCSI})},
  pages = {1-7},
  year = {2022},
  organization = {IEEE}
}

@article{Shields2023,
  author = {Emily Shields and Masudul Imtiaz},
  title = {A Review of Recent Advancements in Sensors Employed in Unmanned Underwater Vehicles},
  year = {2023},
  journal = {Preprints.org}
}

@inproceedings{Hidalgo2015,
  author = {Franco Hidalgo and Thomas Bräunl},
  title = {Review of Underwater SLAM Techniques},
  booktitle = {Proceedings of the 6th International Conference on Automation, Robotics and Applications},
  year = {2015}
}

@article{Panda2020,
  author = {Madhusmita Panda and Bikramaditya Das and Bidyadhar Subudhi and Bibhuti Bhusan Pati},
  title = {A Comprehensive Review of Path Planning Algorithms for Autonomous Underwater Vehicles},
  journal = {International Journal of Automation and Computing},
  volume = {17},
  number = {3},
  pages = {321-352},
  year = {2020},
  publisher = {Springer}
}

@inproceedings{Yin2021,
  author = {Yin, J and Wang, Y and  Lv, J and Ma, J},
  title = {Study on Underwater Simultaneous Localization and Mapping Based on Different Sensors},
  booktitle = {2021 IEEE 10th Data Driven Control and Learning Systems Conference ({DDCLS})},
  pages = {1-5},
  year = {2021},
  organization = {IEEE}
}

@inproceedings{Petillot1998,
  title={Underwater vehicle path planning using a multi-beam forward looking sonar},
  author={Petillot, Y and Ruiz, I Tena and Lane, DM and Wang, Y and Trucco, E and Pican, N},
  booktitle={IEEE Oceanic Engineering Society. OCEANS'98. Conference Proceedings (Cat. No. 98CH36259)},
  volume={2},
  pages={1194--1199},
  year={1998},
  organization={IEEE}
}

@inproceedings{Sahoo2022,
  author = {Avilash Sahoo and Santosha K. Dwivedy and P. S. Robi},
  title = {Adaptive Neuro Fuzzy {PID} Controller for A Compact Autonomous Underwater Vehicle},
  booktitle = {OCEANS 2022, Hampton Roads},
  pages = {1-5},
  year = {2022},
  organization = {IEEE}
}

@article{Orłowski2022,
  author = {Mateusz Orłowski},
  title = {Directions of Development of the Autonomous Unmanned Underwater Vehicles. A Review},
  journal = {Maritime Technical Journal},
  volume = {1},
  number = {224},
  year = {2022},
  publisher = {sciendo}
}

@inproceedings{Shetty2021,
  author = {Siddhy Ganesh Shetty and Karpagavalli Subramanian},
  title = {Performance Comparison of Controllers for the Autonomous Underwater Vehicle {REMUS 100}},
  booktitle = {2021 7th International Conference on Control, Automation and Robotics ({ICCAR})},
  pages = {1-6},
  year = {2021},
  organization = {IEEE}
}

@inproceedings{Raju2020,
  author = {S. Srinivasulu Raju and G. N. Swamy and Y. Bharath and CH. Naga Nandini},
  title = {Simulation and Performance Analysis of Autonomous Underwater Vehicle using Advanced Control Algorithms},
  booktitle = {International Conference on Communication and Signal Processing},
  pages = {1-8},
  year = {2020},
  organization = {IEEE}
}

@article{Bashir2023,
  author = {Adeel Bashir and Sikandar Khan and Naveed Iqbal and Salem Bashmal and Sami Ullah and Fayyaz and Muhammad Usman},
  title = {A Review of the Various Control Algorithms for Trajectory Control of Unmanned Underwater Vehicles},
  journal = {Sustainability},
  volume = {15},
  number = {14691},
  year = {2023},
  publisher = {MDPI}
}

@article{Danovaro2014,
  title={Challenging the paradigms of deep-sea ecology},
  author={Danovaro, Roberto and Snelgrove, Paul VR and Tyler, Paul},
  journal={Trends in ecology \& evolution},
  volume={29},
  number={8},
  pages={465--475},
  year={2014},
  publisher={Elsevier}
}

@article{Koslow2007,
  title={The silent deep: the discovery, ecology, and conservation of the deep sea},
  author={Koslow, Julian Anthony},
  journal={Oceanography},
  volume={23},
  number={1},
  pages={228},
  year={2007}
}

@book{Artiola2004,
  title={Environmental monitoring and characterization},
  author={Artiola, Janick F and Pepper, Ian L and Brusseau, Mark L},
  year={2004},
  publisher={Academic Press}
}

@article{Lovett2007,
  title={Who needs environmental monitoring?},
  author={Lovett, Gary M and Burns, Douglas A and Driscoll, Charles T and Jenkins, Jennifer C and Mitchell, Myron J and Rustad, Lindsey and Shanley, James B and Likens, Gene E and Haeuber, Richard},
  journal={Frontiers in Ecology and the Environment},
  volume={5},
  number={5},
  pages={253--260},
  year={2007},
  publisher={Wiley Online Library}
}

@inproceedings{Livingston2002,
  title={An augmented reality system for military operations in urban terrain},
  author={Livingston, Mark A and Rosenblum, Lawrence J and Julier, Simon J and Brown, Dennis and Baillot, Yohan and Swan, J Edward and Gabbard, Joseph L and Hix, Deborah and others},
  booktitle={Interservice/Industry Training, Simulation, and Education Conference},
  volume={89},
  year={2002},
  organization={Citeseer}
}

@book{Jaiswal2012,
  title={Military operations research: quantitative decision making},
  author={Jaiswal, Narendar Kumar},
  volume={5},
  year={2012},
  publisher={Springer Science \& Business Media}
}

@article{Hollinger2013,
  title={Active planning for underwater inspection and the benefit of adaptivity},
  author={Hollinger, Geoffrey A and Englot, Brendan and Hover, Franz S and Mitra, Urbashi and Sukhatme, Gaurav S},
  journal={The International Journal of Robotics Research},
  volume={32},
  number={1},
  pages={3--18},
  year={2013},
  publisher={SAGE Publications Sage UK: London, England}
}

@inproceedings{Hollinger2017,
  title={Active classification: Theory and application to underwater inspection},
  author={Hollinger, Geoffrey A and Mitra, Urbashi and Sukhatme, Gaurav S},
  booktitle={Robotics Research: The 15th International Symposium {ISRR}},
  pages={95--110},
  year={2017},
  organization={Springer}
}

@article{Martin2016,
  title={Towards integrated autonomous underwater operations for ocean mapping and monitoring},
  author={Ludvigsen, Martin and S{\o}rensen, Asgeir J},
  journal={Annual Reviews in Control},
  pages={145--157},
  year={2016},
  publisher={Elsevier}
}

@article{Tan2011,
  title={A survey of techniques and challenges in underwater localization},
  author={Tan, Hwee-Pink and Diamant, Roee and Seah, Winston KG and Waldmeyer, Marc},
  journal={Ocean Engineering},
  volume={38},
  number={14-15},
  pages={1663--1676},
  year={2011},
  publisher={Elsevier}
}

@inproceedings{Thale2020,
  title={{ROS} based {SLAM} implementation for Autonomous navigation using Turtlebot},
  author={Thale, Sumegh Pramod and Prabhu, Mihir Mangesh and Thakur, Pranjali Vinod and Kadam, Pratik},
  booktitle={ITM Web of conferences},
  volume={32},
  pages={01011},
  year={2020},
  organization={EDP Sciences}
}

@article{Bijjahalli2020,
  title={Advances in intelligent and autonomous navigation systems for small {UAV}},
  author={Bijjahalli, Suraj and Sabatini, Roberto and Gardi, Alessandro},
  journal={Progress in Aerospace Sciences},
  volume={115},
  pages={100617},
  year={2020},
  publisher={Elsevier}
}

@article{Song2003,
  title={Modeling and simulation of autonomous underwater vehicles: design and implementation},
  author={Song, Feijun and An, P Edgar and Folleco, Andres},
  journal={IEEE journal of Oceanic Engineering},
  volume={28},
  number={2},
  pages={283--296},
  year={2003},
  publisher={IEEE}
}

@article{Eickstedt2010,
  title={The backseat control architecture for autonomous robotic vehicles: A case study with the {IVER 2 AUV}},
  author={Eickstedt, Donald P and Sideleau, Scott R},
  journal={Marine technology society journal},
  volume={44},
  number={4},
  pages={42--54},
  year={2010},
  publisher={Marine Technology Society}
}

@article{Qin2022,
  title={A survey on visual navigation and positioning for autonomous {AUV}},
  author={Qin, Jiangying and Li, Ming and Li, Deren and Zhong, Jiageng and Yang, Ke},
  journal={Remote Sensing},
  volume={14},
  number={15},
  pages={3794},
  year={2022},
  publisher={MDPI}
}

@inproceedings{Eustice2007,
  title={Experimental results in synchronous-clock one-way-travel-time acoustic navigation for autonomous underwater vehicles},
  author={Eustice, Ryan M and Whitcomb, Louis L and Singh, Hanumant and Grund, Matthew},
  booktitle={Proceedings 2007 IEEE International Conference on Robotics and Automation},
  pages={4257--4264},
  year={2007},
  organization={IEEE}
}

@article{Hyakudome2011,
  title={Design of autonomous underwater vehicle},
  author={Hyakudome, Tadahiro},
  journal={International Journal of Advanced Robotic Systems},
  volume={8},
  number={1},
  pages={9},
  year={2011},
  publisher={SAGE Publications Sage UK: London, England}
}

@article{Elkins2010,
  title={The Autonomous Maritime Navigation ({AMN}) project: Field tests, autonomous and cooperative behaviors, data fusion, sensors, and vehicles},
  author={Elkins, Les and Sellers, Drew and Monach, W Reynolds},
  journal={Journal of Field Robotics},
  volume={27},
  number={6},
  pages={790--818},
  year={2010},
  publisher={Wiley Online Library}
}

@article{Yuanzhi2024,
  title={Standard datasets for autonomous navigation and mapping: A full-stack construction methodology},
  author={Liu, Yuanzhi and Fu, Yujia and Qin, Minghui and Xu, Yufeng and Cui, Bin and Liu, Kunhua and Chen, Fengdong and Tao, Wei and Vlaminck, Michiel and Goossens, Bart and others},
  journal={IEEE Transactions on Intelligent Vehicles},
  year={2024},
  publisher={IEEE}
}

@inproceedings{Goodin2024,
  title={The NATURE autonomy stack: an open-source stack for off-road navigation},
  author={Goodin, Christopher and Moore, Marc N and Carruth, Daniel W and Hudson, Christopher R and Cagle, Lucas D and Wapnick, Stefan and Jayakumar, Paramsothy},
  booktitle={Unmanned Systems Technology {XXVI}},
  volume={13055},
  pages={8--17},
  year={2024},
  organization={SPIE}
}

@article{McConnell2022,
  title={Perception for underwater robots},
  author={McConnell, John and Collado-Gonzalez, Ivana and Englot, Brendan},
  journal={Current Robotics Reports},
  volume={3},
  number={4},
  pages={177--186},
  year={2022},
  publisher={Springer}
}

@article{Zhao2005,
  title={Experimental study on advanced underwater robot control},
  author={Zhao, Side and Yuh, Junku},
  journal={IEEE transactions on robotics},
  volume={21},
  number={4},
  pages={695--703},
  year={2005},
  publisher={IEEE}
}

@article{Huang2020,
  title={A review on underwater autonomous environmental perception and target grasp, the challenge of robotic organism capture},
  author={Huang, Hai and Tang, Qirong and Li, Jiyong and Zhang, Wanli and Bao, Xuan and Zhu, Haitao and Wang, Gang},
  journal={Ocean Engineering},
  volume={195},
  year={2020},
  publisher={Elsevier}
}

@article{Chang2022,
  title={An active perception framework for autonomous underwater vehicle navigation under sensor constraints},
  author={Chang, Dongsik and Johnson-Roberson, Matthew and Sun, Jing},
  journal={IEEE Transactions on Control Systems Technology},
  volume={30},
  number={6},
  pages={2301--2316},
  year={2022},
  publisher={IEEE}
}

@article{Zhou2023,
  title={Underwater camera: Improving visual perception via adaptive dark pixel prior and color correction},
  author={Zhou, Jingchun and Liu, Qian and Jiang, Qiuping and Ren, Wenqi and Lam, Kin-Man and Zhang, Weishi},
  journal={International Journal of Computer Vision},
  pages={1--19},
  year={2023},
  publisher={Springer}
}

@article{Yilmaz2008,
  author = {Namik Kemal Yilmaz and Constantinos Evangelinos and Pierre F. J. Lermusiaux and Nicholas M. Patrikalakis},
  title = {Path Planning of Autonomous Underwater Vehicles for Adaptive Sampling Using Mixed Integer Linear Programming},
  journal = {IEEE Journal of Oceanic Engineering},
  volume = {33},
  number = {4},
  pages = {522-537},
  year = {2008},
  publisher = {IEEE}
}

@article{Li2019,
  author = {Daoliang Li and Peng Wang and Ling Du},
  title = {Path Planning Technologies for Autonomous Underwater Vehicles-A Review},
  journal = {IEEE Access},
  year = {2019},
  publisher = {IEEE}
}

@article{Zeng2015,
  author = {Zheng Zeng and Lian Lian and Karl Sammut and Fangpo He and Youhong Tang and Andrew Lammas},
  title = {A survey on path planning for persistent autonomy of autonomous underwater vehicles},
  journal = {Ocean Engineering},
  volume = {110},
  pages = {303-313},
  year = {2015},
  publisher = {Elsevier}
}

@article{Pêtrès2007,
  author = {Clément Pêtrès and Yan Pailhas and Pedro Patrón and Yvan Petillot and Jonathan Evans and David Lane},
  title = {Path Planning for Autonomous Underwater Vehicles},
  journal = {IEEE Transactions on Robotics},
  volume = {23},
  number = {2},
  pages = {331-341},
  year = {2007},
  publisher = {IEEE}
}

@article{Guo2021,
  author = {Yinjing Guo and Hui Liu and Xiaojing Fan and Wenhong Lyu},
  title = {Research Progress of Path Planning Methods for Autonomous Underwater Vehicle},
  journal = {Mathematical Problems in Engineering},
  volume = {2021},
  year = {2021},
  publisher = {Hindawi}
}

@ARTICLE{Pailhas2007,
  author={Clément Pêtrès and Yan Pailhas and Pedro Patrón and Yvan Petillot and Jonathan Evans and David Lane},
  journal={IEEE Transactions on Robotics},
  title={Path Planning for Autonomous Underwater Vehicles},
  year={2007},
  volume={23},
  number={2},
  pages={331-341},
  ISSN={1552-3098},
  month={April}
}

@ARTICLE{Meng2025,
  author={Wenlong Meng and Yanbo Pu and Hang Yu and Ya Gong and Dianhui Chu},
  journal={Ocean Engineering},
  title={A fusion framework with spatial perception and energy-aware strategies for {AUV} path planning in uncharted marine environments},
  year={2025},
  volume={328},
  pages={121095}
}

@ARTICLE{RONGHAO2025,
  author={Ronghao Li and Bingying Zhang and Di Lin},
  journal={IEEE Access},
  title={Emperor Yu Tames the Flood: Water Surface Garbage Cleaning Robot Using Improved {A*} Algorithm in Dynamic Environments},
  year={2025},
  volume={},
  number={},
  pages={}
}

@ARTICLE{QinYuan2023,
  author={QinYuan He and HuaPeng Yu and YuChen Fang},
  journal={Gyroscopy and Navigation},
  title={Deep Learning-Based Inertial Navigation Technology for Autonomous Underwater Vehicle Long-Distance Navigation—A Review},
  year={2023},
  volume={14},
  number={3},
  pages={267-275}
}

@ARTICLE{Ma2023,
  title={A review of terrain aided navigation for underwater vehicles},
  author={Ma, Teng and Ding, Shuoshuo and Li, Ye and Fan, Jiajia},
  journal={Ocean Engineering},
  volume={281},
  pages={114779},
  year={2023},
  publisher={Elsevier}
}

@INCOLLECTION{Petillot2021,
  author={Yvan R. Petillot and Gianluca Antonelli and Giuseppe Casalino and Fausto Ferreira},
  title={Underwater Robots: From Remotely Operated Vehicles to Intervention-Autonomous Underwater Vehicles},
  booktitle={Springer Handbook of Ocean Engineering},
  editor={Manohar Singh},
  year={2021},
  pages={1299-1326}
}

@ARTICLE{OKEREKE2023,
  author={Chinonso E. Okereke, Mohd Murtadha Mohamad},
  journal={IEEE Access},
  title={An Overview of Machine Learning Techniques in Local Path Planning for Autonomous Underwater Vehicles},
  year={2023},
  volume={},
  number={},
  pages={},
}

@ARTICLE{Mingyue2022,
  author={Mingyue Cheng and Quansheng Guan and Fei Ji and Julian Cheng and Yankun Chen},
  journal={IEEE Internet of Things Journal},
  title={Dynamic-Detection-Based Trajectory Planning for Autonomous Underwater Vehicle to Collect Data From Underwater Sensors},
  year={2022},
  volume={9},
  number={15},
  pages={13168-13178},
  doi={10.1109/IOTJ.2021.3094887}
}

@ARTICLE{RAFA2024,
  author={Rafal Kot and Piotr Szymak and Pawel Piskur and Krzysztof Naus},
  journal={IEEE Access},
  title={A Comparative Study of Different Collision Avoidance Systems With Local Path Planning for Autonomous Underwater Vehicles},
  year={2024},
  volume={},
  number={},
  pages={}
}

@INPROCEEDINGS{Hernández2015,
  author={Juan David Hernández and Eduard Vidal and Guillem Vallicrosa and Enric Galceran and Marc Carreras},
  booktitle={2015 IEEE International Conference on Robotics and Automation ({ICRA})},
  title={Online Path Planning for Autonomous Underwater Vehicles in Unknown Environments},
  year={2015},
  pages={2539-2544},
  doi={10.1109/ICRA.2015.7139554}
}

@INPROCEEDINGS{Liam2010,
  author={Liam Paull and Sajad Saeedi and Howard Li and Vincent Myers},
  booktitle={6th annual IEEE Conference on Automation Science and Engineering},
  title={An Information Gain Based Adaptive Path Planning Method for an Autonomous Underwater Vehicle using Sidescan Sonar},
  year={2010},
  pages={1-6},
  doi={10.1109/CASE.2010.5584985}
}

@ARTICLE{Yvan2001,
  author={Yvan Petillot and Ioseba Tena Ruiz and David M. Lane},
  journal={IEEE Journal of Oceanic Engineering},
  title={Underwater Vehicle Obstacle Avoidance and Path Planning Using a Multi-Beam Forward Looking Sonar},
  year={2001},
  volume={26},
  number={2},
  pages={240-251},
  doi={10.1109/48.922795}
}

@ARTICLE{Marin2018,
  author={Pablo Marin-Plaza and Ahmed Hussein and David Martin and Arturo de la Escalera},
  journal={Journal of Advanced Transportation},
  title={Global and Local Path Planning Study in a {ROS}-Based Research Platform for Autonomous Vehicles},
  year={2018},
  volume={2018},
  pages={1-10}
}

@ARTICLE{Zhiqiang2022,
  author={Zhiqiang Jian and Shitao Chen and Songyi Zhang and Yu Chen and Nanning Zheng},
  journal={IEEE Transactions on Intelligent Transportation Systems},
  title={Multi-Model-Based Local Path Planning Methodology for Autonomous Driving: An Integrated Framework},
  year={2022},
  volume={23},
  number={5},
  pages={4187-4199}
}

@INPROCEEDINGS{Santhakumar2008,
  author={Santhakumar. M},
  booktitle={2008 10th Intl. Conf. on Control, Automation, Robotics and Vision},
  title={Coupled, Non-linear Control System Design for Autonomous Underwater Vehicle ({AUV})},
  year={2008},
  pages={},
  doi={}
}

@ARTICLE{Behnaz2022,
  author={Behnaz Hadi and Alireza Khosravi and Pouria Sarhadi},
  journal={Applied Ocean Research},
  title={Deep reinforcement learning for adaptive path planning and control of an autonomous underwater vehicle},
  year={2022},
  volume={129},
}

@ARTICLE{McMahon2016,
  author={James McMahon and Erion Plaku},
  journal={IEEE Journal of Oceanic Engineering},
  title={Mission and Motion Planning for Autonomous Underwater Vehicles Operating in Spatially and Temporally Complex Environments},
  year={2016},
  volume={41},
  number={4},
  pages={893-907}
}

@ARTICLE{Alamdari2021,
  author={Shahab Heshmati-Alamdari and Alexandros Nikou and Dimos V. Dimarogonas},
  journal={IEEE Transactions on Automation Science and Engineering},
  title={Robust Trajectory Tracking Control for Underactuated Autonomous Underwater Vehicles in Uncertain Environments},
  year={2021},
  volume={18},
  number={3},
  pages={1288-1301}
}

@ARTICLE{Heshmati2020,
  author={Shahab Heshmati-Alamdari and George C. Karras and Panos Marantos and Kostas J. Kyriakopoulos},
  journal={IEEE Transactions on Control Systems Technology},
  title={A Robust Predictive Control Approach for Underwater Robotic Vehicles},
  year={2020},
  volume={28},
  number={6},
  pages={2352-2367}
}

@ARTICLE{Zhouhua2019,
  author={Zhouhua Peng and Jiasen Wang and Jun Wang},
  journal={IEEE Transactions on Industrial Electronics},
  title={Constrained Control of Autonomous Underwater Vehicles Based on Command Optimization and Disturbance Estimation},
  year={2019},
  volume={66},
  number={5},
  pages={3627-3637}
}

@article{Morgan2022,
  title={Autonomous underwater manipulation: Current trends in dynamics, control, planning, perception, and future directions},
  author={Morgan, Edward and Carlucho, Ignacio and Ard, William and Barbalata, Corina},
  journal={Current Robotics Reports},
  volume={3},
  number={4},
  pages={187--198},
  year={2022},
  publisher={Springer}
}

@article{Junzhi2021,
  title={A survey of underwater multi-robot systems},
  author={Zhou, Ziye and Liu, Jincun and Yu, Junzhi},
  journal={IEEE/CAA Journal of Automatica Sinica},
  volume={9},
  number={1},
  pages={1--18},
  year={2021},
  publisher={IEEE}
}

@article{Guo2020,
  title={Intelligent collaborative navigation and control for {AUV} tracking},
  author={Guo, Jia and Li, Dongyu and He, Bo},
  journal={IEEE Transactions on Industrial Informatics},
  volume={17},
  number={3},
  pages={1732--1741},
  year={2020},
  publisher={IEEE}
}

@article{Pairet2021,
  title={Online mapping and motion planning under uncertainty for safe navigation in unknown environments},
  author={Pairet, {\`E}ric and Hern{\'a}ndez, Juan David and Carreras, Marc and Petillot, Yvan and Lahijanian, Morteza},
  journal={IEEE Transactions on Automation Science and Engineering},
  volume={19},
  number={4},
  pages={3356--3378},
  year={2021},
  publisher={IEEE}
}

@article{Kennedy2007,
  title={Decoupled modelling and controller design for the hybrid autonomous underwater vehicle: {MACO}},
  author={Kennedy, J. and Gamroth, E. and Bradley, C. and Proctor, A. A. and Heard, G. J.},
  journal={International Journal of the Society for Underwater Technology},
  volume={27},
  number={1},
  pages={11--21},
  year={2007},
  publisher={Society for Underwater Technology}
}

@article{Han2020,
  title={Modeling and Fuzzy Decoupling Control of an Underwater Vehicle-Manipulator System},
  author={Han, Han and Wei, Yanhui and Ye, Xiufen and Liu, Wenzhi},
  journal={IEEE Access},
  volume={8},
  pages={18962--18983},
  year={2020},
  publisher={IEEE}
}

@inproceedings{Volpi2018,
  title={Decoupled Sampling-Based Motion Planning for Multiple Autonomous Marine Vehicles},
  author={Volpi, Nicola Catenacci and Smith, Sim{\'o}n C. and Pascoal, Ant{\'o}nio M. and Simetti, Enrico and Turetta, Alessio and Alibani, Michael and Polani, Daniel},
  booktitle={2018 IEEE International Conference on Robotics and Automation ({ICRA})},
  year={2018},
  organization={IEEE}
}

@article{Edward2022,
  title={Autonomous Underwater Manipulation: Current Trends in Dynamics, Control, Planning, Perception, and Future Directions},
  author={Morgan, Edward and Carlucho, Ignacio and Ard, William and Barbalata, Corina},
  journal={Current Robotics Reports},
  volume={3},
  pages={187--198},
  year={2022},
  publisher={Springer}
}

@article{Niankai2022,
  title={Energy-Optimal Control for Autonomous Underwater Vehicles Using Economic Model Predictive Control},
  author={Yang, Niankai and Chang, Dongsik and Johnson-Roberson, Matthew and Sun, Jing},
  journal={IEEE Transactions on Control Systems Technology},
  volume={30},
  number={6},
  pages={2377--2390},
  year={2022},
  publisher={IEEE}
}

@article{Lawrance2019,
  title={Shared autonomy for low-cost underwater vehicles},
  author={Lawrance, Nicholas and DeBortoli, Robert and Jones, Dylan and McCammon, Seth and Milliken, Lauren and Nicolai, Austin and Somers, Thane and Hollinger, Geoffrey},
  journal={Journal of Field Robotics},
  volume={36},
  number={3},
  pages={495--516},
  year={2019},
  publisher={Wiley Online Library}
}

@inproceedings{Chrpa2015,
  title={On Mixed-Initiative Planning and Control for Autonomous Underwater Vehicles},
  author={Chrpa, Luk{\'a}{\v{s}} and Pinto, Jos{\'e} and Ribeiro, Manuel A. },
  booktitle={2015 IEEE/RSJ International Conference on Intelligent Robots and Systems ({IROS})},
  pages={1685--1690},
  year={2015},
  organization={IEEE}
}

@article{Pineda2018,
  title={The Predictive Functional Control and the Management of Constraints in {GUANAY II} Autonomous Underwater Vehicle Actuators},
  author={Pineda Mu{\~n}oz, Wilman Alonso and Sellier, Alain Gauthier and Gomariz Castro, Spartacus},
  journal={IEEE Access},
  volume={6},
  pages={22353--22367},
  year={2018},
  publisher={IEEE}
}

@article{Hasan2024,
  title={Oceanic Challenges to Technological Solutions: A Review of Autonomous Underwater Vehicle Path Technologies in Biomimicry, Control, Navigation and Sensing},
  author={Hasan, Kayes and Ahmad, Shameem and Liaf, Abrar Fahim and Karimi, Mazaher and Ahmed, Tofael and Shawon, Mehedi Azad and Mekhilef, Saad},
  journal={IEEE Access},
  year={2024},
  publisher={IEEE}
}

@article{Immas2022,
  title={Guidance, navigation, and control of {AUV} for permanent underwater optical networks},
  author={Immas, Alexandre and Alam, Mohammad-Reza},
  journal={IEEE Journal of Oceanic Engineering},
  volume={48},
  number={1},
  pages={43--58},
  year={2022},
  publisher={IEEE}
}

@article{Zhongrui2025,
  title={Autonomous Navigation of Mobile Robots: A Hierarchical Planning--Control Framework with Integrated {DWA and MPC}},
  author={Wang, Zhongrui and Wang, Shuting and Xie, Yuanlong and Xiong, Tifan and Wang, Chao},
  journal={Sensors (Basel, Switzerland)},
  volume={25},
  number={7},
  pages={2014},
  year={2025}
}

@article{Aguado2021,
  title={Functional self-awareness and metacontrol for underwater robot autonomy},
  author={Aguado, Esther and Milosevic, Zorana and Hern{\'a}ndez, Carlos and Sanz, Ricardo and Garzon, Mario and Bozhinoski, Darko and Rossi, Claudio},
  journal={Sensors},
  volume={21},
  number={4},
  pages={1210},
  year={2021},
  publisher={MDPI}
}

@article{Fenucci2024,
  title={A multi-platform Guidance, Navigation and Control system for the autosub family of Autonomous Underwater Vehicles},
  author={Fenucci, Davide and Fanelli, Francesco and Consensi, Alberto and Salavasidis, Georgios and Pebody, Miles and Phillips, Alexander B},
  journal={Control Engineering Practice},
  volume={146},
  pages={105902},
  year={2024},
  publisher={Elsevier}
}

@article{Karthik2024,
title = {Guaranteed safe navigation via state-constraints induced by feedback control},
journal = {Mechatronics},
volume = {102},
pages = {103221},
year = {2024},
issn = {0957-4158},
author = {J. Veejay Karthik and Leena Vachhani},
}

@INPROCEEDINGS{Karthik2021,
  author={Karthik, J Veejay and  Arunkumar, GK and Thomas, Maria and Vachhani, Leena},
  booktitle={2021 Seventh Indian Control Conference ({ICC})}, 
  title={Mobile Robot Navigation using State-Constrained Sliding Mode Control}, 
  year={2021},
  volume={},
  number={},
  pages={219-224},
  doi={10.1109/ICC54714.2021.9703148}}

@article{Mashhood2024,
	year = 2024,
	month = {December},
	publisher = {Preprints},
	author = {Mashhood Zafar and Selahattin Ozcelik},
	title = {Comparative Evaluation of {PID, MPC} and Fuzzy Logic Control Strategies for Energy Transfer Stations in District Cooling Networks},
	journal = {Preprints}
}

@article{Salimzhan2015,
title = {Autonomous Unmanned Underwater Vehicles Development Tendencies},
journal = {Procedia Engineering},
volume = {106},
pages = {141-148},
year = {2015},
note = {Proceedings of the 2nd International Conference on Dynamics and Vibroacoustics of Machines (DVM2014) September 15 –17, 2014 Samara, Russia},
issn = {1877-7058},
author = {Salimzhan A. Gafurov and Evgeniy V. Klochkov}
}

@inproceedings{Chen2023,
author = {Chen, Tuochao and Chan, Justin and Gollakota, Shyamnath},
year = {2023},
title = {Underwater {3D} positioning on smart devices},
isbn = {9798400702365},
publisher = {Association for Computing Machinery},
address = {New York, NY, USA},
booktitle = {Proceedings of the {ACM SIGCOMM} 2023 Conference},
pages = {33–48},
numpages = {16},
location = {New York, NY, USA},
series = {ACM SIGCOMM}
}

@article{Sun2024,
  title={Underwater robots and key technologies for operation control},
  author={Sun, Linxiang and Wang, Yu and Hui, Xiaolong and Ma, Xibo and Bai, Xuejian and Tan, Min},
  journal={Cyborg and Bionic Systems},
  volume={5},
  pages={0089},
  year={2024},
  publisher={AAAS}
}

@article{Russell2024,
  title={Evaluation of Technological Readiness in Mixed Maturity Sub-systems of Large Uncrewed Underwater Vehicles},
  author={Yu, Ning and Russell, Sebastian and Flawith, James and Nicholas, Nicholas},
  journal={The {ITEA} Journal of Test and Evaluation},
  volume={45},
  number={4},
  year={2024}
}

@article{Murai2025,
  author = {Murai, Swapnil and Vairagi, Rahul Das and Semwal, Vijay},
  title = {Optimal Trajectory Generation of Various English Alphabets Using Deep Learning Model for {3-R} Manipulator},
  journal = {Journal of Field Robotics},
  pages = {},
  year = {2025},
  eprint = {https://onlinelibrary.wiley.com/doi/pdf/10.1002/rob.22537}
}

@inproceedings{gomes2018,
  title={A model predictive control scheme for autonomous underwater vehicle formation control},
  author={Gomes, Rui and Pereira, Fernando Lobo},
  booktitle={2018 13th APCA International Conference on Automatic Control and Soft Computing (CONTROLO)},
  pages={195--200},
  year={2018},
  organization={IEEE}
}

\nocite{*}



\end{document}